\documentclass[10pt]{article}

\usepackage{microtype}
\usepackage{times}
\usepackage{subcaption}
\usepackage{graphicx}
\usepackage{amsmath}
\usepackage[round]{natbib}
\usepackage{geometry}
\usepackage{setspace}
\usepackage{titlesec}
\usepackage{float}
\usepackage{placeins}
\usepackage{xcolor}
\usepackage{tikz}
\usepackage{titling}
\usepackage{makecell}
\usepackage[colorlinks=true,linkcolor=blue,citecolor=blue,urlcolor=blue]{hyperref}
\usepackage{hypcap}
\usetikzlibrary{backgrounds}

\newcommand{\alphaValue}{0.5}

\newcommand{\colorword}[4]{%
    \begingroup%
    \definecolor{tempcolor}{RGB}{#1,#2,#3}%
    \tikz[baseline=(text.base)]{%
        \node[inner sep=0pt, outer sep=0pt, anchor=base west] (text) {\strut #4};
        \begin{scope}[on background layer]
            \fill [tempcolor, opacity=\alphaValue] (text.south west) rectangle (text.north east);
        \end{scope}
    }%
    \endgroup%
}

\titleformat{\section}[block]{\bfseries\Large}{\thesection}{1em}{}
\titleformat{\subsection}[block]{\bfseries\large}{\thesubsection}{1em}{}

\makeatletter
\renewenvironment{abstract}{
  \if@twocolumn
    \section*{\abstractname}
  \else
    \begin{center}
      {\Large\bfseries \abstractname\vspace{\z@}}
    \end{center}
    \quotation
  \fi}
{\if@twocolumn\else\endquotation\fi}
\makeatother

\pretitle{\vspace{-30pt}\begin{center}\LARGE\hrule height 2pt\vspace{20pt}}
\posttitle{\vspace{20pt}\hrule height 2pt\end{center}\vspace{20pt}}

\title{\textbf{Detecting Memorization in Large Language Models}}
\author{\textbf{Eduardo Slonski}\thanks{Correspondence to eduardoslonski@gmail.com}}

\date{}

\begin{document}

\maketitle

\begin{abstract}
Large language models (LLMs) have achieved impressive results in natural language processing but are prone to memorizing portions of their training data, which can compromise evaluation metrics, raise privacy concerns, and limit generalization. Traditional methods for detecting memorization rely on output probabilities or loss functions, often lacking precision due to confounding factors like common language patterns. In this paper, we introduce an analytical method that precisely detects memorization by examining neuron activations within the LLM. By identifying specific activation patterns that differentiate between memorized and not memorized tokens, we train classification probes that achieve near-perfect accuracy. The approach can also be applied to other mechanisms, such as repetition, as demonstrated in this study, highlighting its versatility. Intervening on these activations allows us to suppress memorization without degrading overall performance, enhancing evaluation integrity by ensuring metrics reflect genuine generalization. Additionally, our method supports large-scale labeling of tokens and sequences, crucial for next-generation AI models, improving training efficiency and results. Our findings contribute to model interpretability and offer practical tools for analyzing and controlling internal mechanisms in LLMs.
\end{abstract}

\section{Introduction}
\label{sec:introduction}

Large Language Models (LLMs) have revolutionized natural language processing by demonstrating unprecedented abilities in text generation, comprehension, and a variety of applications. These models excel in tasks ranging from machine translation and summarization to creative writing and complex problem-solving, largely due to their extensive training on vast and diverse datasets. However, alongside these impressive capabilities, LLMs exhibit a propensity to memorize segments of their training data verbatim, which can lead to overfitting, privacy concerns, and challenges in evaluation.

Memorization within LLMs presents a double-edged sword. While it allows models to recall specific facts or phrases essential for certain tasks, excessive memorization can damage their ability to generalize to new challenges and may result in the unintended disclosure of sensitive information from the training data. Moreover, verbatim reproduction complicates the evaluation of models, as it can inflate performance metrics without reflecting genuine understanding or reasoning capabilities.

Previous approaches to detecting memorization often rely on examining output probabilities or loss functions \citep{carlini2020extracting}, under the assumption that memorized tokens lead to highly confident predictions with near-zero loss. While intuitive, these methods struggle with confounding factors like common phrases and predictable patterns that generate similar outputs. They also often lack the precision and interpretability needed for in-depth analysis and intervention.

In this paper, we propose an analytical method that detects memorization in LLMs with near-perfect accuracy by focusing on the model's internal mechanisms. Our approach involves analyzing neuron activations to distinguish between memorized and not memorized tokens. By identifying specific activation patterns that effectively separate the two groups, we train classification probes capable of detecting memorization with close to 100\% accuracy.

Furthermore, we demonstrate that intervening on these activations allows us to suppress memorization and repetition mechanisms, altering the model's behavior without degrading overall performance.

By leveraging a curated and diverse dataset to differentiate between memorized and not memorized sequences, we can label millions of tokens in a large and general dataset. This approach combines the benefits of both targeted analysis and broad applicability, addressing a gap where many existing methods focus on one aspect or the other.

Our method offers significant advantages for large-scale token and sequence labeling pipelines, which are crucial for next-generation AI. This scalable approach enables systematic improvement of training data quality and nuanced evaluation of model behaviors, opening new possibilities for targeted optimization.

\subsection*{Our main contributions are as follows:}

\begin{enumerate}
    \item \textbf{Precise Detection of Memorization:} We introduce a methodology that leverages neuron activations within the LLM and classification probes to accurately detect memorized sequences. Our analysis reveals that certain activation patterns are highly indicative of memorization, enabling precise classification.
    \item \textbf{Mechanism-Focused Approach:} By concentrating on the internal mechanisms of the model rather than solely on output behaviors, we achieve greater interpretability. This allows us to understand how memorization manifests within the model's architecture and to distinguish it from other phenomena.
    \item \textbf{Applicability to Other Mechanisms:} We demonstrate that our method is not limited to memorization. By applying the same analytical framework, we successfully detect other mechanisms such as repetition, achieving similarly high accuracy. This highlights the versatility of our approach in probing and understanding various internal processes of LLMs.
    \item \textbf{Enhancing Evaluation Integrity:} Our method enables the detection of memorization during model evaluation, ensuring that performance metrics genuinely reflect the model's capability to generalize rather than its ability to recall training data. This is critical for developing reliable benchmarks and advancing the field.
    \item \textbf{Intervention Capability:} Leveraging the insights from our analysis, we show that it is possible to intervene in the model's activations to alter its behavior. Specifically, we can suppress the memorization and repetition mechanisms, compelling the model to rely on alternative processes for generating predictions.
\end{enumerate}

Our findings enhance model interpretability and offer practical tools for analyzing and controlling internal mechanisms in large language models. By establishing a framework for understanding neural mechanisms, our methodology paves the way for more sophisticated methods of model analysis and control.

Beyond memorization detection, our approach provides a broader framework for understanding LLMs' internal operations. As models advance and become increasingly sophisticated, tools to understand and direct their inner workings become ever more crucial. Identifying and intervening on specific activation patterns opens new possibilities for studying other mechanisms within LLMs, such as reasoning and knowledge integration.

An illustrative example of our method's effectiveness is presented in \autoref{fig:figure1}, where we visualize sequences processed by the memorization probe. The not memorized sequences are depicted in red, while the memorized sequences are shown in blue. This clear differentiation underscores the precision of our approach in identifying memorized content within the model.

\begin{figure}[H]
    \capstart
    \begin{minipage}{\textwidth}
    \begin{flushleft}
    {\footnotesize 
    \selectfont
    \colorword{255}{101}{101}{Ind}%
    \colorword{255}{30}{30}{ifference}%
    \colorword{255}{0}{0}{,}%
    \colorword{255}{0}{0}{ after}%
    \colorword{255}{0}{0}{ all}%
    \colorword{255}{0}{0}{,}%
    \colorword{255}{3}{3}{ is}%
    \colorword{255}{0}{0}{ more}%
    \colorword{255}{0}{0}{ dangerous}%
    \colorword{255}{12}{12}{ than}%
    \colorword{255}{4}{4}{ anger}%
    \colorword{255}{1}{1}{ and}%
    \colorword{255}{0}{0}{ hatred}%
    \colorword{255}{2}{2}{.}%
    \colorword{255}{29}{29}{ An}%
    \colorword{255}{6}{6}{ger}%
    \colorword{255}{0}{0}{ can}%
    \colorword{255}{17}{17}{ at}%
    \colorword{255}{15}{15}{ times}%
    \colorword{255}{0}{0}{ be}%
    \colorword{255}{6}{6}{ creative}%
    \colorword{255}{12}{12}{.}%
    \colorword{255}{15}{15}{ One}%
    \colorword{255}{12}{12}{ writes}%
    \colorword{255}{14}{14}{ a}%
    \colorword{255}{33}{33}{ great}%
    \colorword{255}{8}{8}{ poem}%
    
    \colorword{255}{17}{17}{,}%
    \colorword{255}{16}{16}{ a}%
    \colorword{255}{33}{33}{ great}%
    \colorword{255}{52}{52}{ sym}%
    \colorword{255}{21}{21}{phony}%
    \colorword{255}{10}{10}{.}%
    \colorword{255}{16}{16}{ One}%
    \colorword{255}{26}{26}{ does}%
    \colorword{255}{3}{3}{ something}%
    \colorword{255}{0}{0}{ special}%
    \colorword{255}{14}{14}{ for}%
    \colorword{255}{6}{6}{ the}%
    \colorword{255}{0}{0}{ sake}%
    \colorword{255}{0}{0}{ of}%
    \colorword{255}{0}{0}{ humanity}%
    \colorword{255}{2}{2}{ because}%
    \colorword{255}{21}{21}{ one}%
    \colorword{255}{0}{0}{ is}%
    \colorword{255}{0}{0}{ angry}%
    \colorword{255}{0}{0}{ at}%
    \colorword{255}{0}{0}{ the}%
    \colorword{255}{0}{0}{ injustice}%
    \colorword{255}{12}{12}{ that}%
    
    \colorword{255}{6}{6}{ one}%
    \colorword{255}{6}{6}{ witnesses}%
    \colorword{255}{6}{6}{.}%
    \colorword{255}{14}{14}{ But}%
    \colorword{255}{4}{4}{ indifference}%
    \colorword{255}{0}{0}{ is}%
    \colorword{255}{19}{19}{ never}%
    \colorword{255}{9}{9}{ creative}%
    \colorword{255}{12}{12}{.}%
    \colorword{255}{123}{123}{\textbackslash n}%
    
    \colorword{255}{14}{14}{\textbackslash n}%
    
    \colorword{255}{25}{25}{It}%
    \colorword{255}{3}{3}{'s}%
    \colorword{255}{7}{7}{ liberty}%
    \colorword{255}{0}{0}{ or}%
    \colorword{255}{36}{36}{ it}%
    \colorword{255}{18}{18}{'s}%
    \colorword{255}{9}{9}{ death}%
    \colorword{255}{1}{1}{.}%
    \colorword{255}{15}{15}{ It}%
    \colorword{255}{6}{6}{'s}%
    \colorword{255}{0}{0}{ freedom}%
    \colorword{255}{0}{0}{ for}%
    \colorword{255}{0}{0}{ everybody}%
    \colorword{255}{5}{5}{ or}%
    \colorword{255}{0}{0}{ freedom}%
    \colorword{255}{13}{13}{ for}%
    \colorword{255}{10}{10}{ nobody}%
    \colorword{255}{0}{0}{.}%
    \colorword{255}{18}{18}{ America}%
    \colorword{255}{3}{3}{ today}%
    \colorword{255}{0}{0}{ finds}%
    \colorword{255}{4}{4}{ herself}%
    \colorword{255}{15}{15}{ in}%
    \colorword{255}{9}{9}{ a}%
    \colorword{255}{5}{5}{ unique}%
    
    \colorword{255}{8}{8}{ situation}%
    \colorword{255}{0}{0}{.}%
    \colorword{255}{0}{0}{ Histor}%
    \colorword{255}{0}{0}{ically}%
    \colorword{255}{2}{2}{,}%
    \colorword{255}{106}{106}{ rev}%
    \colorword{255}{10}{10}{olutions}%
    \colorword{255}{5}{5}{ are}%
    \colorword{255}{7}{7}{ bloody}%
    \colorword{255}{0}{0}{.}%
    \colorword{255}{25}{25}{ Oh}%
    \colorword{255}{21}{21}{ yes}%
    \colorword{255}{0}{0}{,}%
    \colorword{255}{33}{33}{ they}%
    \colorword{255}{0}{0}{ are}%
    \colorword{255}{0}{0}{.}%
    \colorword{255}{22}{22}{ They}%
    \colorword{255}{21}{21}{ haven}%
    \colorword{255}{32}{32}{'t}%
    \colorword{255}{44}{44}{ never}%
    \colorword{255}{26}{26}{ had}%
    \colorword{255}{33}{33}{ a}%
    \colorword{255}{32}{32}{ blood}%
    \colorword{255}{41}{41}{-}%
    \colorword{255}{4}{4}{less}%
    \colorword{255}{28}{28}{ revolution}%
    \colorword{255}{30}{30}{,}%
    \colorword{255}{1}{1}{ or}%
    \colorword{255}{24}{24}{ a}%
    
    \colorword{255}{6}{6}{ non}%
    \colorword{255}{12}{12}{-}%
    \colorword{255}{19}{19}{violent}%
    \colorword{255}{35}{35}{ revolution}%
    \colorword{255}{14}{14}{.}%
    \colorword{255}{23}{23}{\textbackslash n}%
    
    \colorword{255}{1}{1}{\textbackslash n}%
    
    \colorword{255}{5}{5}{I}%
    \colorword{255}{31}{31}{ have}%
    \colorword{255}{4}{4}{ a}%
    \colorword{255}{82}{82}{ dream}%
    \colorword{0}{0}{255}{ that}%
    \colorword{11}{11}{255}{ my}%
    \colorword{0}{0}{255}{ four}%
    \colorword{9}{9}{255}{ little}%
    \colorword{0}{0}{255}{ children}%
    \colorword{0}{0}{255}{ will}%
    \colorword{0}{0}{255}{ one}%
    \colorword{0}{0}{255}{ day}%
    \colorword{0}{0}{255}{ live}%
    \colorword{0}{0}{255}{ in}%
    \colorword{0}{0}{255}{ a}%
    \colorword{0}{0}{255}{ nation}%
    \colorword{0}{0}{255}{ where}%
    \colorword{0}{0}{255}{ they}%
    \colorword{0}{0}{255}{ will}%
    \colorword{0}{0}{255}{ not}%
    \colorword{0}{0}{255}{ be}%
    \colorword{8}{8}{255}{ judged}%
    \colorword{7}{7}{255}{ by}%
    \colorword{9}{9}{255}{ the}%
    \colorword{12}{12}{255}{ color}%
    \colorword{4}{4}{255}{ of}%
    
    \colorword{17}{17}{255}{ their}%
    \colorword{0}{0}{255}{ skin}%
    \colorword{0}{0}{255}{ but}%
    \colorword{0}{0}{255}{ by}%
    \colorword{0}{0}{255}{ the}%
    \colorword{0}{0}{255}{ content}%
    \colorword{1}{1}{255}{ of}%
    \colorword{0}{0}{255}{ their}%
    \colorword{0}{0}{255}{ character}%
    \colorword{0}{0}{255}{.}%
    \colorword{0}{0}{255}{ I}%
    \colorword{0}{0}{255}{ have}%
    \colorword{0}{0}{255}{ a}%
    \colorword{0}{0}{255}{ dream}%
    \colorword{0}{0}{255}{ today}%
    \colorword{0}{0}{255}{!}%
    \colorword{255}{137}{137}{\textbackslash n}%
    
    \colorword{255}{115}{115}{\textbackslash n}%
    
    \colorword{255}{22}{22}{The}%
    \colorword{255}{70}{70}{ one}%
    \colorword{255}{0}{0}{ thing}%
    \colorword{255}{20}{20}{ I}%
    \colorword{255}{24}{24}{ want}%
    \colorword{255}{140}{140}{ to}%
    \colorword{255}{73}{73}{ make}%
    \colorword{255}{35}{35}{ sure}%
    \colorword{255}{32}{32}{ of}%
    \colorword{255}{11}{11}{ is}%
    \colorword{255}{20}{20}{ that}%
    \colorword{255}{15}{15}{ every}%
    \colorword{255}{8}{8}{ penny}%
    \colorword{255}{14}{14}{ that}%
    \colorword{255}{0}{0}{ is}%
    \colorword{255}{20}{20}{ contributed}%
    \colorword{255}{0}{0}{ to}%
    \colorword{255}{9}{9}{ this}%
    \colorword{255}{1}{1}{ campaign}%
    \colorword{255}{33}{33}{ is}%
    \colorword{255}{0}{0}{ a}%
    \colorword{255}{5}{5}{ matter}%
    \colorword{255}{16}{16}{ of}%
    \colorword{255}{0}{0}{ public}%
    
    \colorword{255}{14}{14}{ record}%
    \colorword{255}{13}{13}{.}%
    \colorword{255}{0}{0}{ And}%
    \colorword{255}{14}{14}{ I}%
    \colorword{255}{6}{6}{'}%
    \colorword{255}{46}{46}{m}%
    \colorword{255}{6}{6}{ proud}%
    \colorword{255}{7}{7}{ of}%
    \colorword{255}{17}{17}{ it}%
    \colorword{255}{9}{9}{.}%
    \colorword{255}{5}{5}{ And}%
    \colorword{255}{1}{1}{ I}%
    \colorword{255}{20}{20}{'}%
    \colorword{255}{15}{15}{m}%
    \colorword{255}{8}{8}{ proud}%
    \colorword{255}{11}{11}{ of}%
    \colorword{255}{0}{0}{ the}%
    \colorword{255}{4}{4}{ fact}%
    \colorword{255}{0}{0}{ that}%
    \colorword{255}{10}{10}{ no}%
    \colorword{255}{0}{0}{ one}%
    \colorword{255}{35}{35}{ can}%
    \colorword{255}{27}{27}{ say}%
    \colorword{255}{16}{16}{ I}%
    \colorword{255}{13}{13}{'}%
    \colorword{255}{63}{63}{ve}%
    \colorword{255}{18}{18}{ ever}%
    \colorword{255}{10}{10}{ done}%
    \colorword{255}{15}{15}{ anything}%
    \colorword{255}{9}{9}{ that}%
    \colorword{255}{7}{7}{ was}%
    \colorword{255}{14}{14}{ financially}%
    
    \colorword{255}{6}{6}{ dishon}%
    \colorword{255}{152}{152}{est}%
    \colorword{255}{0}{0}{ in}%
    \colorword{255}{0}{0}{ my}%
    \colorword{255}{0}{0}{ public}%
    \colorword{255}{0}{0}{ career}%
    \colorword{255}{0}{0}{.}%
    \colorword{255}{10}{10}{\textbackslash n}%
    
    \colorword{255}{17}{17}{\textbackslash n}%
    
    \colorword{255}{0}{0}{Four}%
    \colorword{255}{12}{12}{ score}%
    \colorword{181}{181}{255}{ and}%
    \colorword{255}{186}{186}{ seven}%
    \colorword{255}{120}{120}{ years}%
    \colorword{14}{14}{255}{ ago}%
    \colorword{14}{14}{255}{ our}%
    \colorword{2}{2}{255}{ fathers}%
    \colorword{0}{0}{255}{ brought}%
    \colorword{0}{0}{255}{ forth}%
    \colorword{0}{0}{255}{ on}%
    \colorword{3}{3}{255}{ this}%
    \colorword{1}{1}{255}{ continent}%
    \colorword{0}{0}{255}{,}%
    \colorword{0}{0}{255}{ a}%
    \colorword{0}{0}{255}{ new}%
    \colorword{0}{0}{255}{ nation}%
    \colorword{0}{0}{255}{,}%
    \colorword{0}{0}{255}{ conceived}%
    \colorword{0}{0}{255}{ in}%
    \colorword{0}{0}{255}{ Liberty}%
    \colorword{0}{0}{255}{,}%
    \colorword{0}{0}{255}{ and}%
    
    \colorword{40}{40}{255}{ dedicated}%
    \colorword{2}{2}{255}{ to}%
    \colorword{0}{0}{255}{ the}%
    \colorword{0}{0}{255}{ proposition}%
    \colorword{7}{7}{255}{ that}%
    \colorword{0}{0}{255}{ all}%
    \colorword{0}{0}{255}{ men}%
    \colorword{6}{6}{255}{ are}%
    \colorword{9}{9}{255}{ created}%
    \colorword{16}{16}{255}{ equal}%
    \colorword{1}{1}{255}{.}%
    
    \colorword{255}{255}{255}{}%
    }
    \caption{Visualization of our detection results. The detection ranges from red (not memorized) to blue (memorized). As we can see, the detection is extremely precise, completely separating the memorized sequences from the not memorized ones, even when the sequences are randomly arranged together.}
    \label{fig:figure1}
    \end{flushleft}
    \end{minipage}
\end{figure}

\section{Related Work}
\label{sec:related_work}

Understanding and detecting memorization in large language models have garnered significant attention due to its implications for privacy, generalization, and model interpretability. Previous studies have primarily focused on quantifying and extracting memorized data, as well as exploring the factors that contribute to memorization in neural networks.

\citet{carlini2020extracting} introduced practical attacks to extract verbatim memorized training data from LLMs using black-box query access. They formalized the notion of \textit{k-eidetic memorization} and developed methods for generating and ranking potential memorized samples. Their work highlighted that larger models are more prone to memorization and discussed mitigation strategies such as differential privacy and data de-duplication.

Further exploring the quantification of memorization, \citet{carlini2022quantifying} established that memorization scales log-linearly with model size and data duplication. They demonstrated that longer prompts increase the discoverability of memorized data, emphasizing challenges in auditing and mitigating memorization in large models.

\citet{huang2024demystifying} conducted controlled experiments by injecting specific sequences into training data to study verbatim memorization. They observed that non-trivial repetition is required for memorization and that later training checkpoints memorize more effectively. Their findings suggest that verbatim memorization is intertwined with general language modeling capabilities, making it difficult to suppress without degrading model quality.

\citet{biderman2023emergent} investigated the predictability of memorization behavior in LLMs. They found that smaller or partially trained models are unreliable predictors of memorization in larger models, identifying emergent properties not predictable from smaller scales. Their scaling laws provide insights into forecasting memorization behavior but also highlight the limitations of extrapolation methods.

Efforts to localize memorization within models have been explored by \citet{maini2023can} and \citet{chang2023localize}. \citet{maini2023can} demonstrated that memorization is not confined to specific layers but distributed across neurons scattered throughout the network. They introduced ``example-tied dropout'' to localize memorization to predetermined neurons, effectively mitigating memorization with minimal impact on generalization. \citet{chang2023localize} proposed benchmarks to evaluate localization methods, finding that precise localization of memorization remains challenging due to shared neurons among related sequences.

In terms of probing model internals, \citet{meng2022locating} identified mid-layer feed-forward networks in GPT models as key components for storing factual associations. They introduced causal mediation analysis to trace neuron activations critical for factual predictions and proposed methods to edit model weights for updating specific facts.

Our work differs from these studies by proposing an analytical method that detects memorization with high precision and interpretability. Instead of focusing on external extraction attacks or global quantification, we analyze neuron activations to distinguish between memorized and not memorized tokens. By identifying specific activations that separate the two groups, we train classification probes that achieve near-perfect accuracy in detecting memorization. This approach not only reveals where memorization occurs within the model but also enables interventions to alter the model's behavior, providing a practical tool for understanding and controlling memorization in LLMs.

\section{Detecting Memorization}
\label{sec:detecting_memorization}

Previous studies have attempted to detect when a language model uses memorization to predict the next token by examining the loss function (e.g., \citealp{carlini2020extracting}). This approach is intuitive because memorized tokens usually narrow the output to a single confident prediction, causing the loss to approach zero. However, this method faces challenges since other mechanisms can produce similar effects on the output. For instance, some studies (e.g., \citealp{meng2022locating}) aim to identify where information is stored in large language models (LLMs) by precisely intervening in the forward computation to determine which components affect the result.

In contrast, we propose an analytical method that is highly precise, achieving an accuracy close to 100\%, and interpretable.

\subsection{Methodology}
\label{sec:methodology}

Our approach involves first collecting samples that are memorized by the LLM and comparing their activations with similar, not memorized samples. We then identify neuron activations that best distinguish between the two groups and use these activations to label a larger dataset, which is subsequently used to train classification probes. We use the Pythia 1B model \citep{biderman2023pythia} for this study.

\subsubsection*{Part 1: Identifying Neuron Activations}

\paragraph{1. Gathering Memorized Samples}

We identified several sources likely to be memorized by the LLM, including famous quotes, speeches, Bible passages, legal texts, manuals, poems, pledges, licenses, nursery rhymes, anthems, passages from famous novels, song lyrics, common disclaimers, and more. We manually tested each sample on the LLM and retained those that were memorized, indicated by a very high confidence level on the correct predictions. It is essential to have a small yet sufficiently diverse corpus of samples; our corpus comprised 100 memorized samples.

\paragraph{2. Gathering Not Memorized Samples}

This step is crucial and can lead to incorrect results if not conducted properly. We balanced the not memorized samples with the memorized ones by including a very similar but not memorized sample for each memorized sample. For instance, for a memorized speech, we included a not memorized speech of a similar style and length. It is important to avoid using random samples, as this can introduce bias due to differences in distribution; memorized samples are more likely to originate from the sources cited above.\footnote{We also explored using LLM-generated text for this step, either by prompting a powerful LLM to produce samples with the same format and style as the memorized ones but different content, or by iteratively having an LLM rewrite the memorized sample with changes until it becomes a completely new, unmemorized sample. This technique can be very useful for larger-scale studies.}

\paragraph{3. Labeling the Tokens}
\label{sec:labeling_the_tokens}

We manually labeled the tokens of both the memorized and not memorized groups, paying attention to three key considerations:

\begin{enumerate}
    \item \textbf{Partial Memorization}: Sometimes, not the entire sample is memorized. It is important to include only the tokens that are actually memorized.
    \item \textbf{Tokenization Issues}: Due to tokenization, some words are split into multiple tokens (e.g., ``gira'' and ``ffe'' for ``giraffe''). We only use the last token of a word because the preceding tokens use different mechanisms to predict the next token.
    \item \textbf{Dataset Balance}: To avoid biasing the labeled token dataset toward specific texts, we limit tokens from each sample to 100. For example, using all tokens from the U.S. Constitution would overrepresent that single text.
\end{enumerate}

Our final corpus contains 10,000 tokens, evenly split between memorized and not memorized tokens.

\paragraph{4. Detecting Neuron Activations}
\label{sec:detecting_activations}

We recorded activations for all samples and analyzed them statistically, comparing labeled memorized versus not memorized tokens. Features were ranked using Cohen's $d$ \eqref{eq:cohen_combined}, which measures group separation using pooled standard deviation. Other methods (ROC AUC, Wilcoxon, $t$-test, Kolmogorov-Smirnov, Jensen-Shannon, Wasserstein, energy statistics, Levene's test, kurtosis) yielded similar results for this task. Distribution-focused methods revealed an interesting finding discussed in Section~\ref{sec:interpretability}.

\begin{equation}
Cohen's\; d = \frac{M_1 - M_2}{SD_{\text{pooled}}} \quad \text{where} \quad SD_{\text{pooled}} = \sqrt{\frac{(n_1 - 1)SD_1^2 + (n_2 - 1)SD_2^2}{n_1 + n_2 - 2}}
\label{eq:cohen_combined}
\end{equation}

\noindent
\textit{$M_i$, $SD_i$, and $n_i$ are the means, standard deviations, and sample sizes of groups $i = 1, 2$.}

Our analysis revealed that many neuron activations are related to memorization and can effectively separate the two groups. We consider a Cohen's $d$ value of 1 or greater to be indicative of an effective separation.

\begin{figure}[h]
    \capstart
    \centering
    \includegraphics[width=\textwidth]{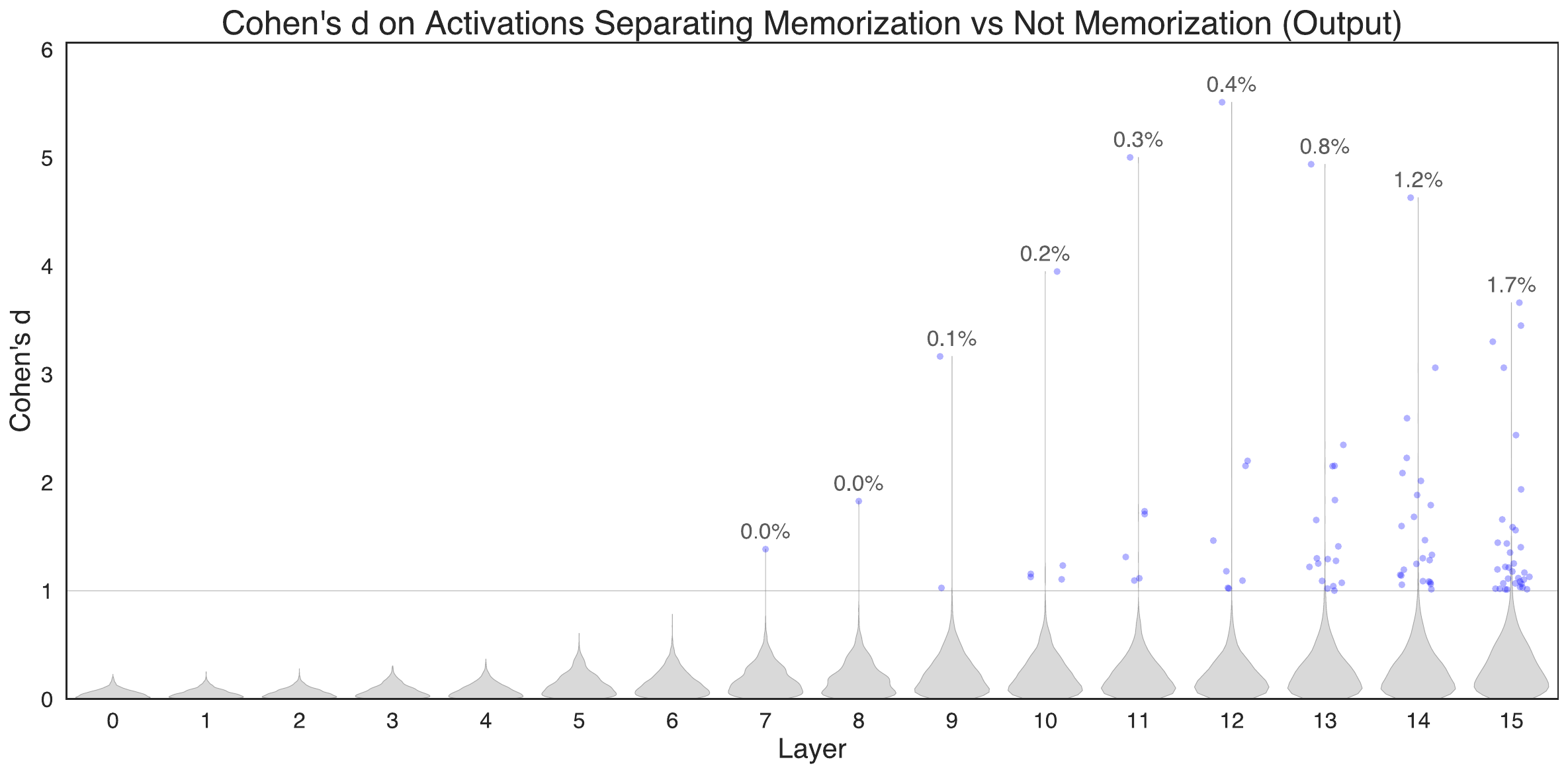}
    \caption{Distribution of Cohen's $d$ Values for Output Activations Separating Memorized vs.\ Not Memorized Tokens. We highlight the activations with a Cohen's $d$ above 1 and indicate their proportion among all activations.}
    \label{fig:figure2}
\end{figure}

In \autoref{fig:figure2}, we show the Cohen's $d$ distribution for output activations across layers. The distributions become taller in later layers, indicating more activations with larger Cohen's $d$ values and greater separation. Notably, there is one specific activation consistently at the top across all layers, activation 1668, which we discuss further in Section~\ref{sec:certainty}.

\begin{figure}[h]
    \capstart
    \centering
    \includegraphics[width=\textwidth]{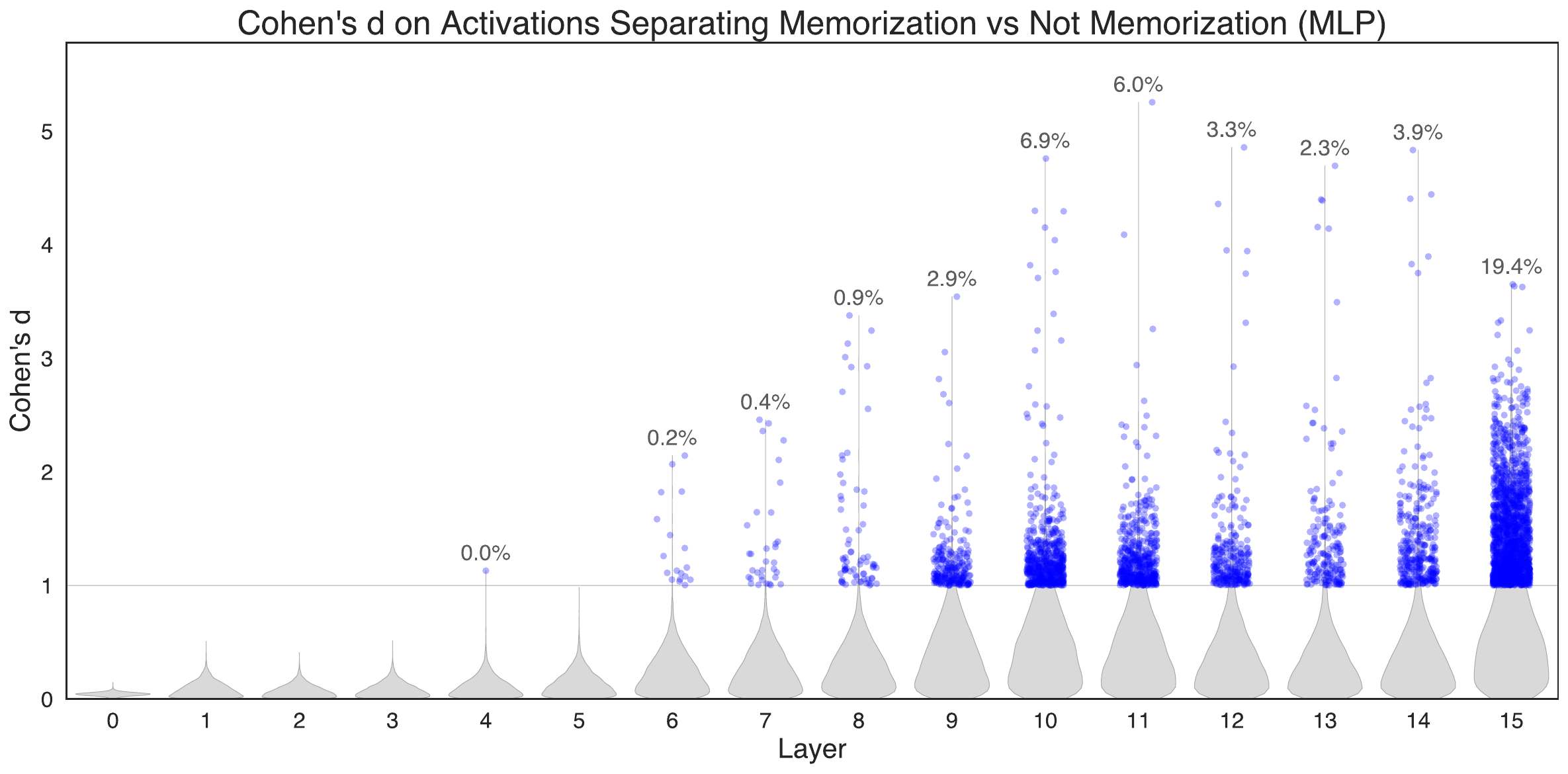}
    \caption{Distribution of Cohen's $d$ Values for MLP Activations Separating Memorized vs.\ Not Memorized Tokens. It is visually clear how much more effectively the MLP activations can separate the two groups compared to other activation types, not only in the number of neurons but also in their proportion among all neurons.}
    \label{fig:figure3}
\end{figure}

We performed this analysis for all activation types in the model, with results available in \autoref{app:violin-plots-memorization}. We present the same plot for the intermediate activations of the MLP (Multilayer Perceptron) in \autoref{fig:figure3}. We selected the intermediate MLP activations because they exhibit the largest proportion of separable activations between memorized and not memorized tokens. This observation aligns with several reasons:

\begin{enumerate}
    \item \textbf{Feature Extraction}: The MLP serves as a strong feature extractor by default.
    \item \textbf{Knowledge Storage}: It has been credited as the primary location of factual knowledge in Transformers \citep{meng2022locating}.
    \item \textbf{Computational Freedom}: It is not part of a skip connection, giving it more freedom to create and utilize features, as it is not directly added to future computations.
\end{enumerate}

\begin{figure}[H]
    \capstart
    \centering
    \includegraphics[width=\textwidth]{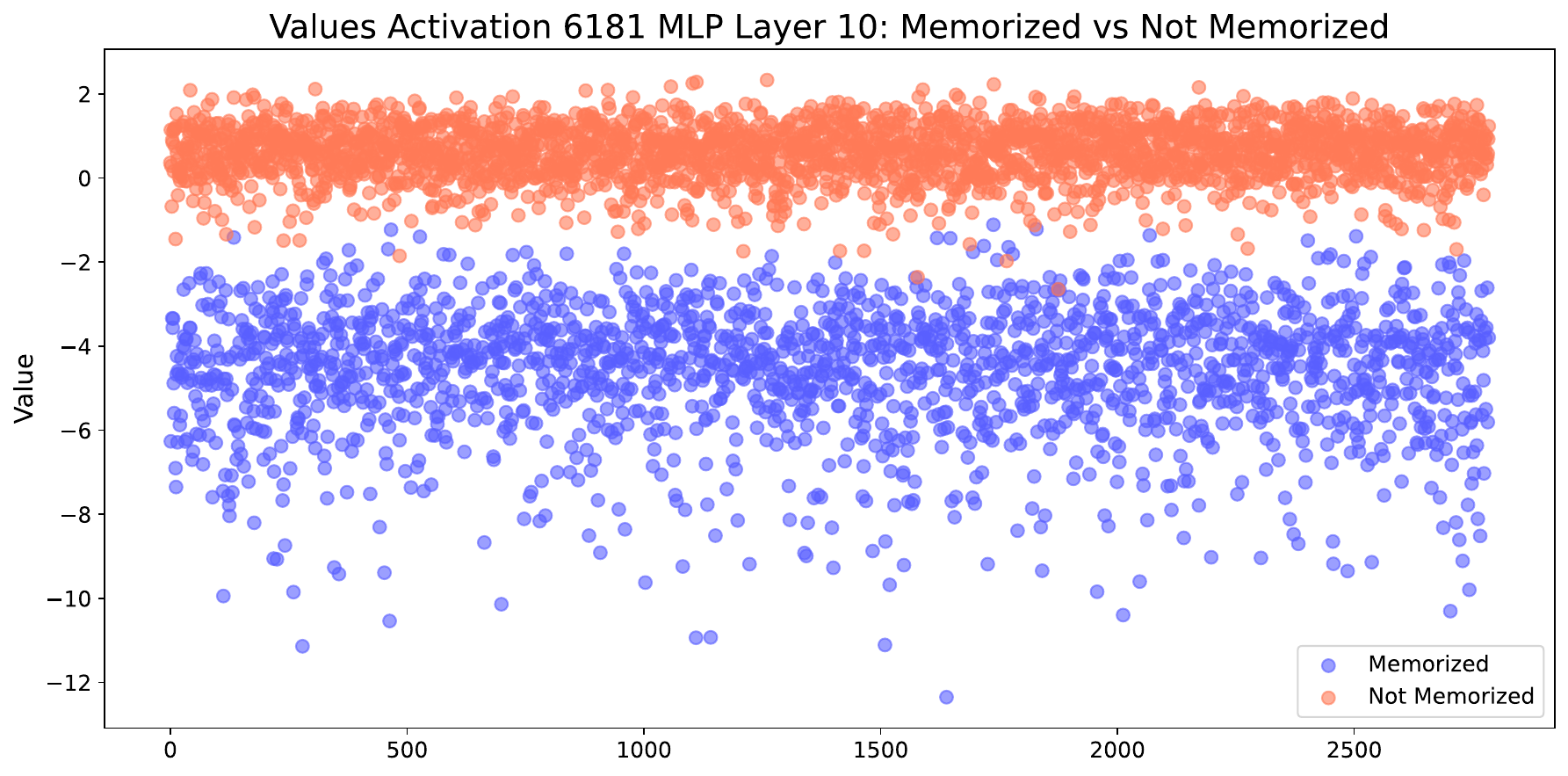}
    \caption{Activation Values for Neuron 6181 in MLP Layer 10. The Memorized (blue) and Not Memorized (orange) are clearly separated.}
    \label{fig:figure4}
\end{figure}

For illustration, \autoref{fig:figure4} shows the activation values for neuron 6181 in the MLP at layer 10, demonstrating how well it separates the memorized and not memorized groups.

\begin{figure}[h]
    \capstart
    \centering
    \includegraphics[width=\textwidth]{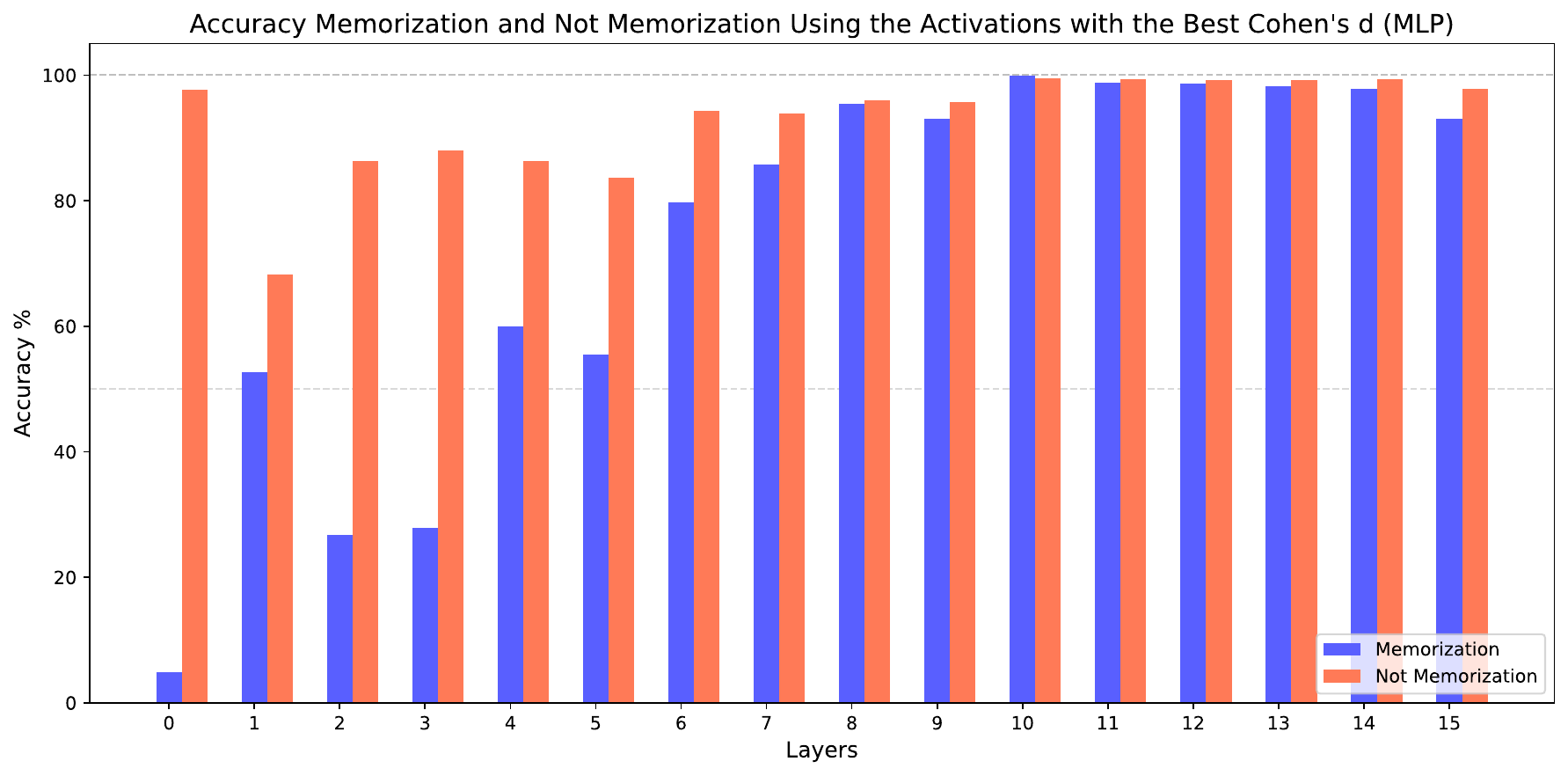}
    \caption{Classification Accuracy Using Best Activation on the MLP per layer.}
    \label{fig:figure5}
\end{figure}

\autoref{fig:figure5} presents the classification accuracy achieved by using the activation with the highest Cohen's $d$ value from the MLP to distinguish between memorized and not memorized tokens.

These results indicate that some individual activations can be used to detect memorized sequences in LLMs. As shown in \autoref{fig:figure5}, the activation at layer 10 can distinguish between memorized and not memorized tokens with 99\% accuracy. This is a significant finding, demonstrating that the model not only develops a specific and robust mechanism for memorization but also that this mechanism is widely utilized within the model's activations and can be readily identified.

Additional plots for all activation types are provided in \autoref{app:classification-accuracy-cohens-d}{}.

As a reference, \autoref{fig:figure6} visualizes the activation values of neuron 6181 in MLP layer 10 for two speeches: a memorized speech (shown in blue) and a not memorized speech (shown in red).

\begin{figure}[h]
    \capstart
    \begin{minipage}{\textwidth}
    \begin{flushleft}
    {\footnotesize
    \selectfont
    \textbf{Memorized Speech:}\\
    \colorword{255}{128}{128}{Four}%
    \colorword{255}{67}{67}{ score}%
    \colorword{245}{245}{255}{ and}%
    \colorword{239}{239}{255}{ seven}%
    \colorword{255}{171}{171}{ years}%
    \colorword{255}{95}{95}{ ago}%
    \colorword{255}{75}{75}{ our}%
    \colorword{102}{102}{255}{ fathers}%
    \colorword{23}{23}{255}{ brought}%
    \colorword{55}{55}{255}{ forth}%
    \colorword{90}{90}{255}{ on}%
    \colorword{44}{44}{255}{ this}%
    \colorword{54}{54}{255}{ continent}%
    \colorword{9}{9}{255}{,}%
    \colorword{0}{0}{255}{ a}%
    \colorword{50}{50}{255}{ new}%
    \colorword{127}{127}{255}{ nation}%
    \colorword{50}{50}{255}{,}%
    \colorword{61}{61}{255}{ conceived}%
    \colorword{109}{109}{255}{ in}%
    \colorword{126}{126}{255}{ Liberty}%
    \colorword{57}{57}{255}{,}%
    \colorword{16}{16}{255}{ and}%
    \colorword{59}{59}{255}{ dedicated}%
    
    \colorword{42}{42}{255}{ to}%
    \colorword{49}{49}{255}{ the}%
    \colorword{118}{118}{255}{ proposition}%
    \colorword{45}{45}{255}{ that}%
    \colorword{57}{57}{255}{ all}%
    \colorword{74}{74}{255}{ men}%
    \colorword{93}{93}{255}{ are}%
    \colorword{172}{172}{255}{ created}%
    \colorword{191}{191}{255}{ equal}%
    \colorword{54}{54}{255}{.}%

    \vspace{1em}
    \textbf{Not Memorized Speech:}\\

    \colorword{255}{129}{129}{At}%
    \colorword{255}{44}{44}{ the}%
    \colorword{255}{33}{33}{ end}%
    \colorword{255}{0}{0}{ of}%
    \colorword{255}{66}{66}{ your}%
    \colorword{255}{74}{74}{ life}%
    \colorword{255}{66}{66}{,}%
    \colorword{255}{41}{41}{ you}%
    \colorword{255}{79}{79}{ will}%
    \colorword{255}{72}{72}{ never}%
    \colorword{255}{62}{62}{ regret}%
    \colorword{255}{44}{44}{ not}%
    \colorword{255}{24}{24}{ having}%
    \colorword{255}{70}{70}{ passed}%
    \colorword{255}{77}{77}{ one}%
    \colorword{255}{122}{122}{ more}%
    \colorword{255}{74}{74}{ test}%
    \colorword{255}{7}{7}{,}%
    \colorword{255}{44}{44}{ winning}%
    \colorword{255}{34}{34}{ one}%
    \colorword{255}{82}{82}{ more}%
    \colorword{255}{59}{59}{ verdict}%
    \colorword{255}{51}{51}{,}%
    \colorword{255}{51}{51}{ or}%
    \colorword{255}{59}{59}{ not}%
    \colorword{255}{81}{81}{ closing}%
    \colorword{255}{59}{59}{ one}%
    
    \colorword{255}{124}{124}{ more}%
    \colorword{255}{82}{82}{ deal}%
    \colorword{255}{17}{17}{.}%
    \colorword{255}{27}{27}{ You}%
    \colorword{255}{44}{44}{ will}%
    \colorword{255}{80}{80}{ regret}%
    \colorword{255}{59}{59}{ time}%
    \colorword{255}{80}{80}{ not}%
    \colorword{255}{61}{61}{ spent}%
    \colorword{255}{51}{51}{ with}%
    \colorword{255}{61}{61}{ a}%
    \colorword{255}{69}{69}{ husband}%
    \colorword{255}{121}{121}{,}%
    \colorword{255}{137}{137}{ a}%
    \colorword{255}{109}{109}{ child}%
    \colorword{255}{106}{106}{,}%
    \colorword{255}{113}{113}{ a}%
    \colorword{255}{83}{83}{ friend}%
    \colorword{255}{74}{74}{,}%
    \colorword{255}{80}{80}{ or}%
    \colorword{255}{102}{102}{ a}%
    \colorword{255}{86}{86}{ parent}%
    \colorword{255}{88}{88}{.}%

    \colorword{255}{255}{255}{}%
    }
    \caption{Visualization of Activation Values for Neuron 6181 in MLP Layer 10. The color scale represents activation values ranging from $-6$ (blue) to $2$ (red), with a threshold at $-1.3$ (white).}
    \label{fig:figure6}
    \end{flushleft}
    \end{minipage}
\end{figure}

Although using individual activations could be a viable solution for detecting memorization, we identify several limitations in relying solely on this approach:

\begin{enumerate}
    \item \textbf{Degrees of Memorization}: While the range of activation values is significant, it does not necessarily correlate directly with the degree to which a token is memorized.
    \item \textbf{Potential Artifacts}: Relying on a feature not explicitly designed to detect memorization may introduce artifacts or unintended biases.
    \item \textbf{Limited Representational Power}: There may exist more effective internal representations that cannot be captured by examining a single or a small group of activations.
\end{enumerate}

To overcome these limitations and enhance our detection capability, we leverage our previous findings to train classification probes.

\subsubsection*{Part 2: Training Classification Probes}

\paragraph{1. Choosing Activations}

We first selected the activations that best separated the two groups, as explained in the previous section. We performed both manual and automated tests on each activation to ensure they were suitable as labelers. We identified several activations capable of performing this task and ultimately chose the one that yielded the most reliable results, activation 1857 in the MLP at layer 10.\footnote{It is important to find neuron activations that are reliable for labeling tokens, not just for separating the groups. We observed that some activations represent other features of which memorization is a part but not the sole component, such as certainty, which we discuss in Section~\ref{sec:certainty}. To address this, we included samples with repetitions, knowledge retrieval, and pattern matching, labeling them as not memorized, to avoid using an activation that captures multiple properties.}

\paragraph{2. Labeling Tokens}
\label{sec:labeling_tokens}

Using the selected activation, we labeled memorized tokens in a larger dataset. We utilized the SlimPajama dataset \citep{soboleva2023slimpajama} and randomly selected a subset of 200,000 samples (approximately 400 million tokens), excluding the GitHub portion due to the distinct patterns found in code compared to general text. We processed the samples through the model and selected memorized and not memorized tokens using the following procedure:

\begin{itemize}
    \item \textbf{Window Size}: We employed a window size of at least 10 tokens, where all tokens had activation values within the memorized threshold.
    \item \textbf{Ignoring Completion Tokens}: We ignored completion tokens, same procedure mentioned in Section~\ref{sec:labeling_the_tokens} Tokenization Issues.
    \item \textbf{Selecting Not Memorized Tokens}: We applied the same procedure to select not memorized tokens by inverting the threshold.
    \item \textbf{Handling Atypical Cases}: We separately saved sequences that met the memorization criteria but had an average cross-entropy loss greater than 2, which is atypical for memorized sequences.
\end{itemize}

We saved all activations for each token and ultimately obtained one million memorized and one million not memorized tokens.

\paragraph{3. Training the Probe}

We trained probe models on each activation type. Specifically, we trained linear probes and two-layer probes with a ReLU non-linear activation in the hidden layer. The probes were trained to classify tokens as memorized (1) or not memorized (0).

\section{Results}
\label{sec:results}

\begin{figure}[h]
    \capstart
    \centering
    \includegraphics[width=0.8\textwidth]{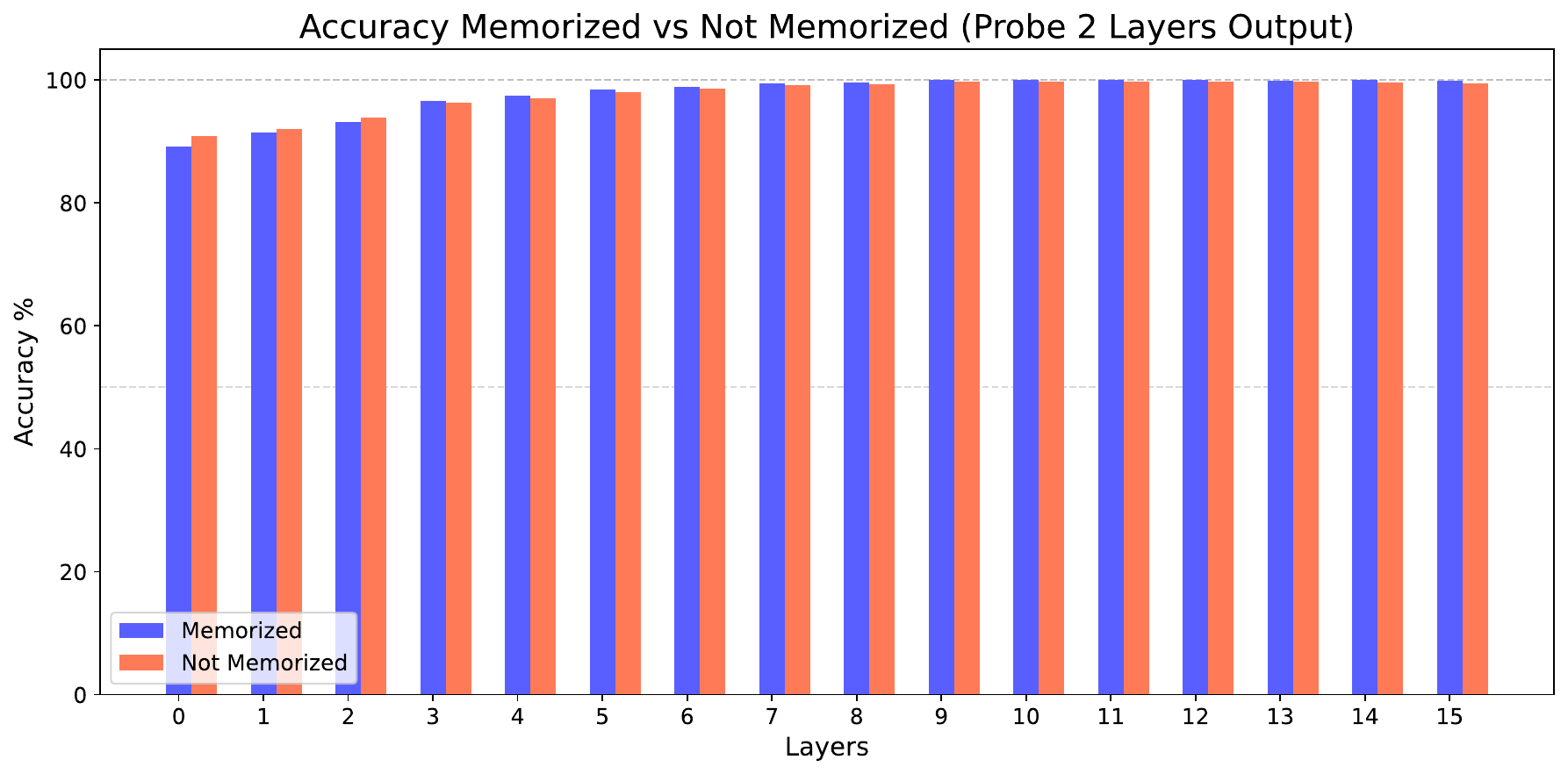}
    \caption{Accuracy Comparison Using Two-Layer Probe on Output Activations}
    \label{fig:figure7}
\end{figure}

The probes were able to classify memorization with 99.9\% accuracy. \autoref{fig:figure7} shows the classification accuracy of memorized and not memorized tokens across the layers in the test set. The accuracy reaches a very high level as early as layer 3 and continues to approach 100\% in subsequent layers.

\begin{figure}[h]
    \capstart
    \centering
    \includegraphics[width=0.8\textwidth]{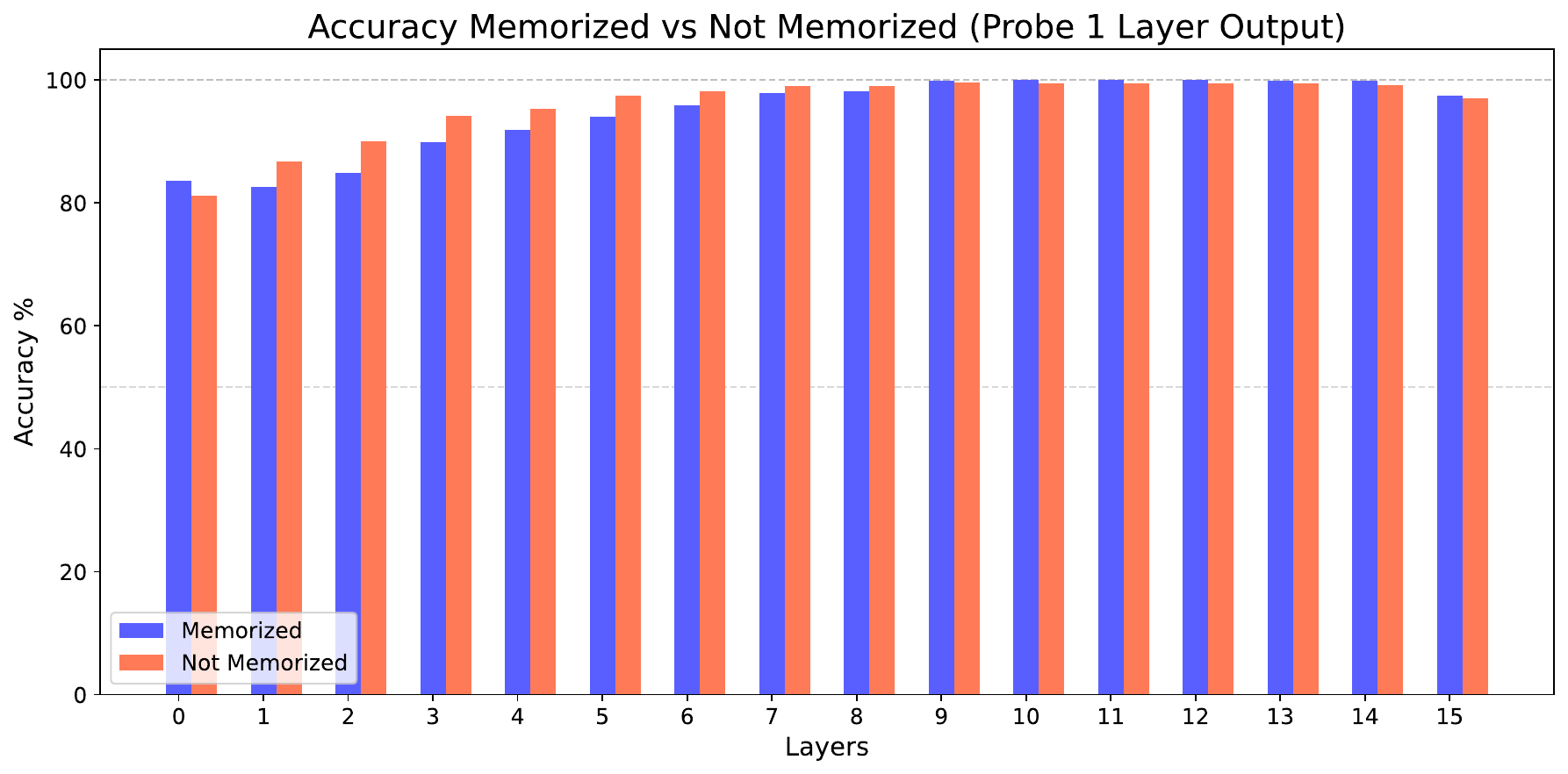}
    \caption{Accuracy Comparison Using One-Layer Probe on Output Activations}
    \label{fig:figure8}
\end{figure}

It is important to note that the plots in Figures~\ref{fig:figure7} and \ref{fig:figure8} show accuracy on a validation set derived from the same dataset used for training the probe. This dataset comprises memorized sentences and randomly selected not memorized sentences. However, this may not be an entirely unbiased test because the memorized sentences originate from specific distributions (e.g., books, speeches, licenses, disclaimers), while the not memorized sentences are randomly selected and can belong to any distribution. Consequently, the probe might exploit distributional differences rather than focusing solely on the memorization feature, potentially lowering its loss by learning these distributional cues. The plots accurately reflect the probe's ability to identify memorized sentences within common LLM training data but may not necessarily indicate its effectiveness in distinguishing a memorized versus a not memorized poem, for example, when both belong to the same distribution. To address this, we evaluated the probe on a curated dataset comprising memorized and not memorized samples from the same distribution.

\begin{figure}[H]
    \capstart
    \centering
    \includegraphics[width=0.8\textwidth]{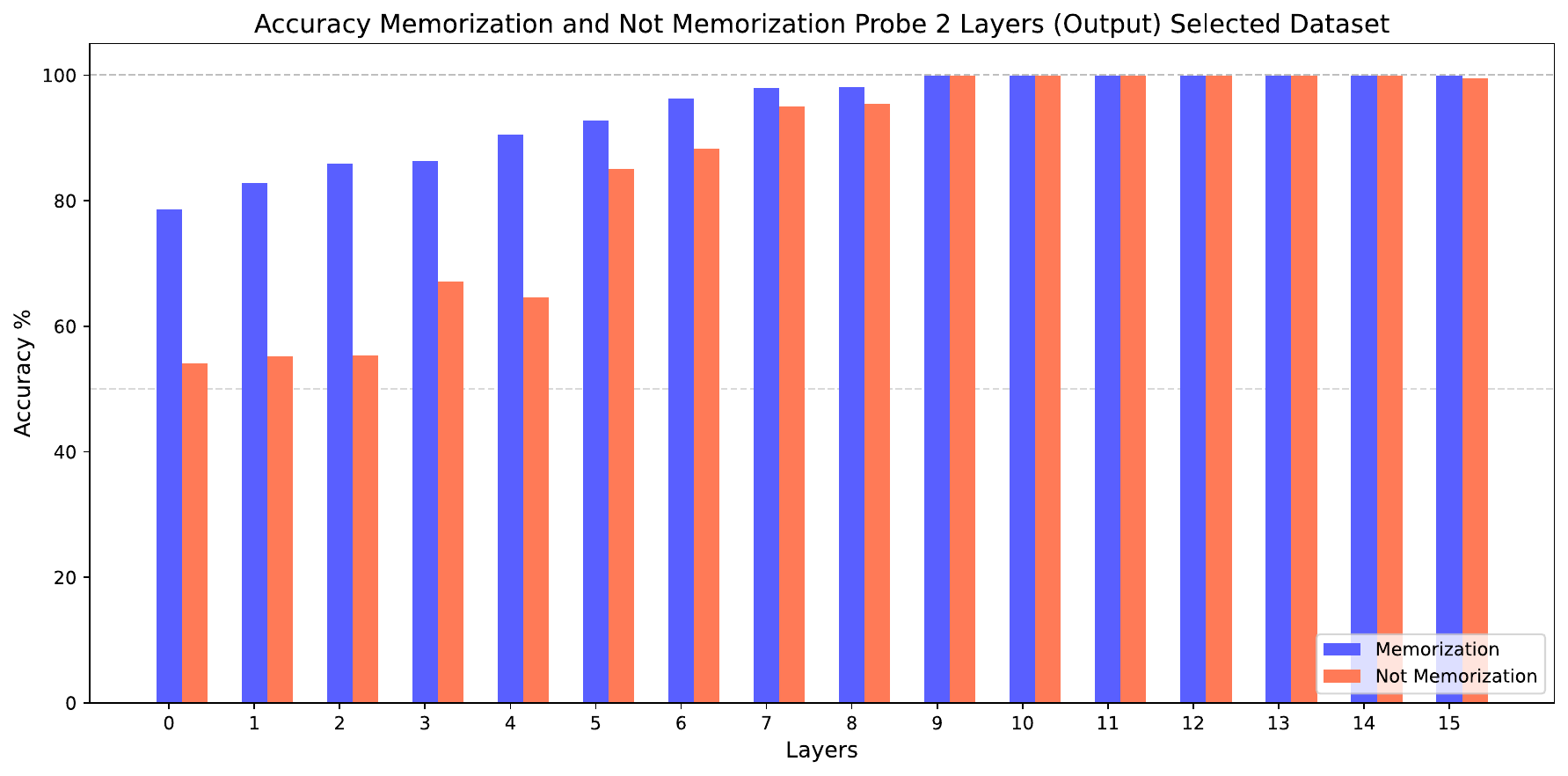}
    \caption{Accuracy of the probe on a curated dataset where memorized and not memorized samples are from the same distribution.}
    \label{fig:figure9}
\end{figure}

\autoref{fig:figure9} presents the classification accuracy on the curated dataset using the two-layer probe.

As we can see, the accuracy remains near 100\% even in the curated dataset, which is balanced and presents a more challenging set of samples. This demonstrates the robustness of our method in detecting memorization across different distributions.

\begin{figure}[h]
    \capstart
    \centering
    \includegraphics[width=0.8\textwidth]{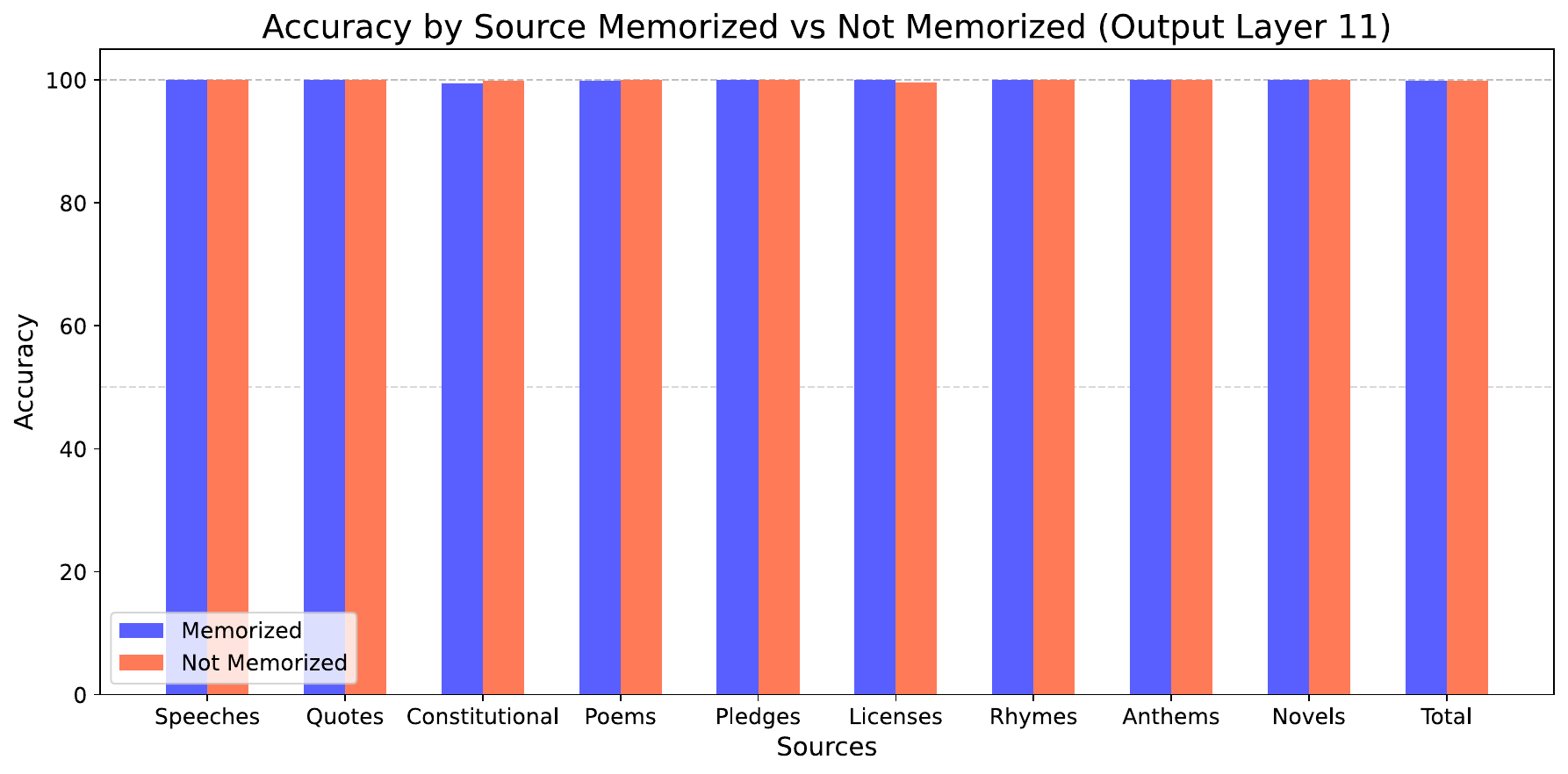}
    \caption{Classification accuracy by data source at Output Layer 11, illustrating the probe's performance across different types of text.}
    \label{fig:figure10}
\end{figure}

\autoref{fig:figure10} shows the classification accuracy by source for memorized and not memorized tokens at output layer 11.

\section{Repetition}
\label{sec:repetition}

\begin{figure}[h]
    \capstart
    \centering
    \includegraphics[width=\textwidth]{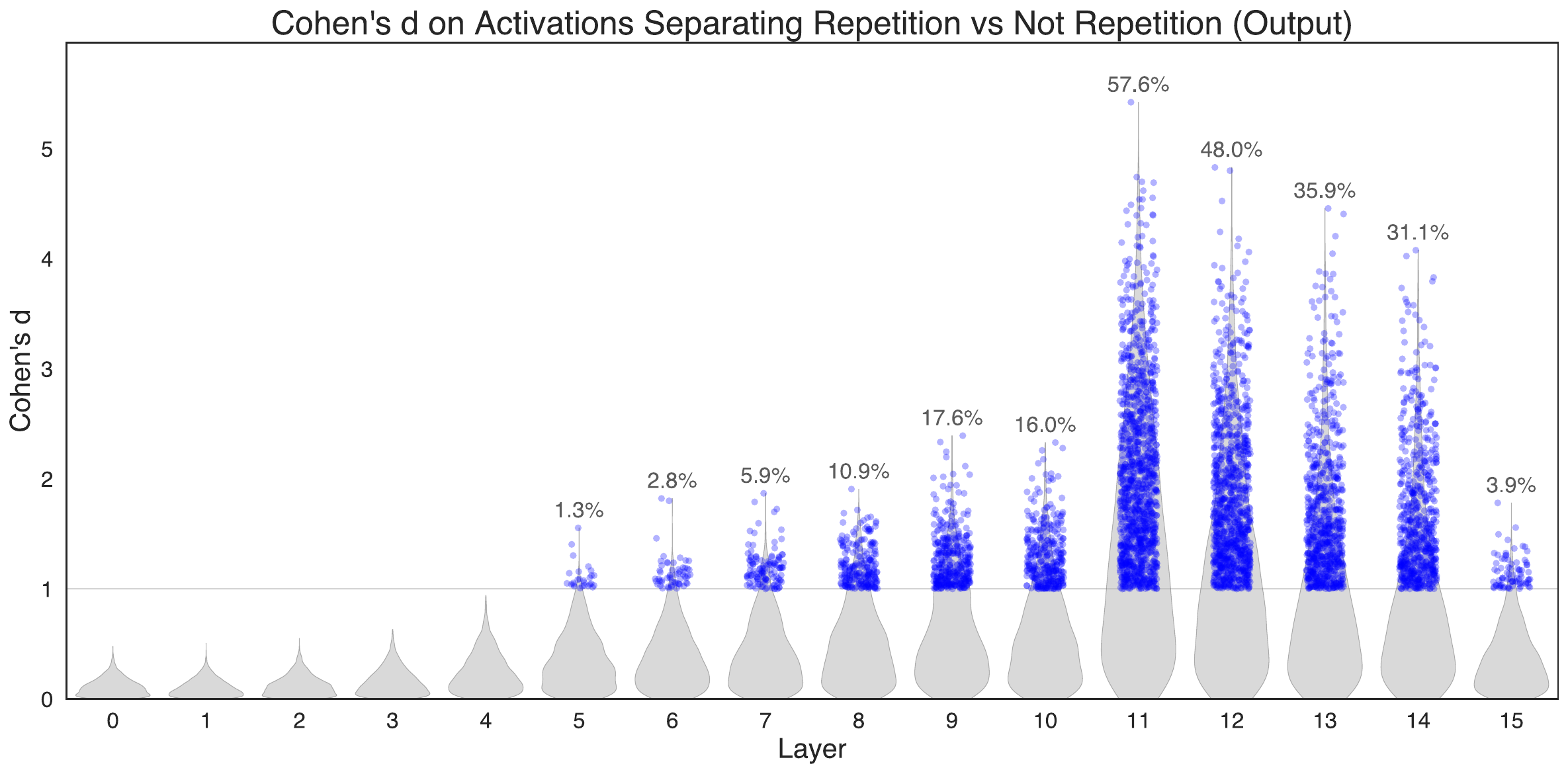}
    \caption{Distribution of Cohen's $d$ Values for Activations Separating Repetition vs.\ Not Repetition (Output). It is interesting to observe the jump in separable activations at Layer 11. We've observed that Layer 11 is very special in terms of transition from context enriching to next token prediction.}
    \label{fig:figure11}
\end{figure}

To demonstrate the applicability of our method to another mechanism, we applied the same approach described in Section~\ref{sec:methodology} to train probes for detecting repetition. Repetition occurs when the model simply copies a sequence that has previously appeared in the text. As shown in \autoref{fig:figure11}, we present the distribution of activations that effectively separate repeated from not repeated text.

\begin{figure}[H]
    \capstart
    \centering
    \includegraphics[width=0.8\textwidth]{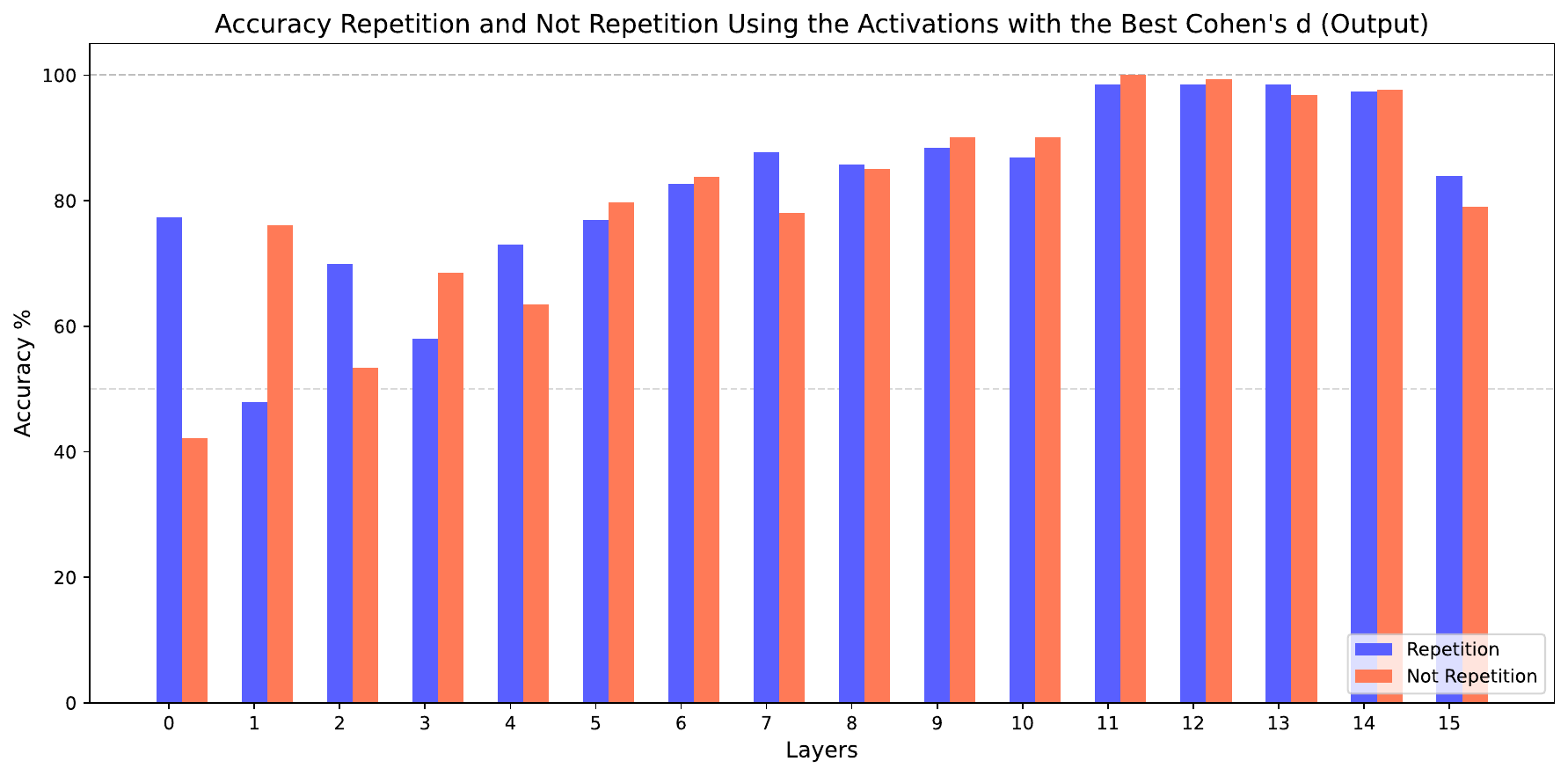}
    \caption{Classification Accuracy Using Best Activation on the MLP per layer for Repetition Detection}
    \label{fig:figure12}
\end{figure}

\begin{figure}[h]
    \capstart
    \centering
    \includegraphics[width=0.8\textwidth]{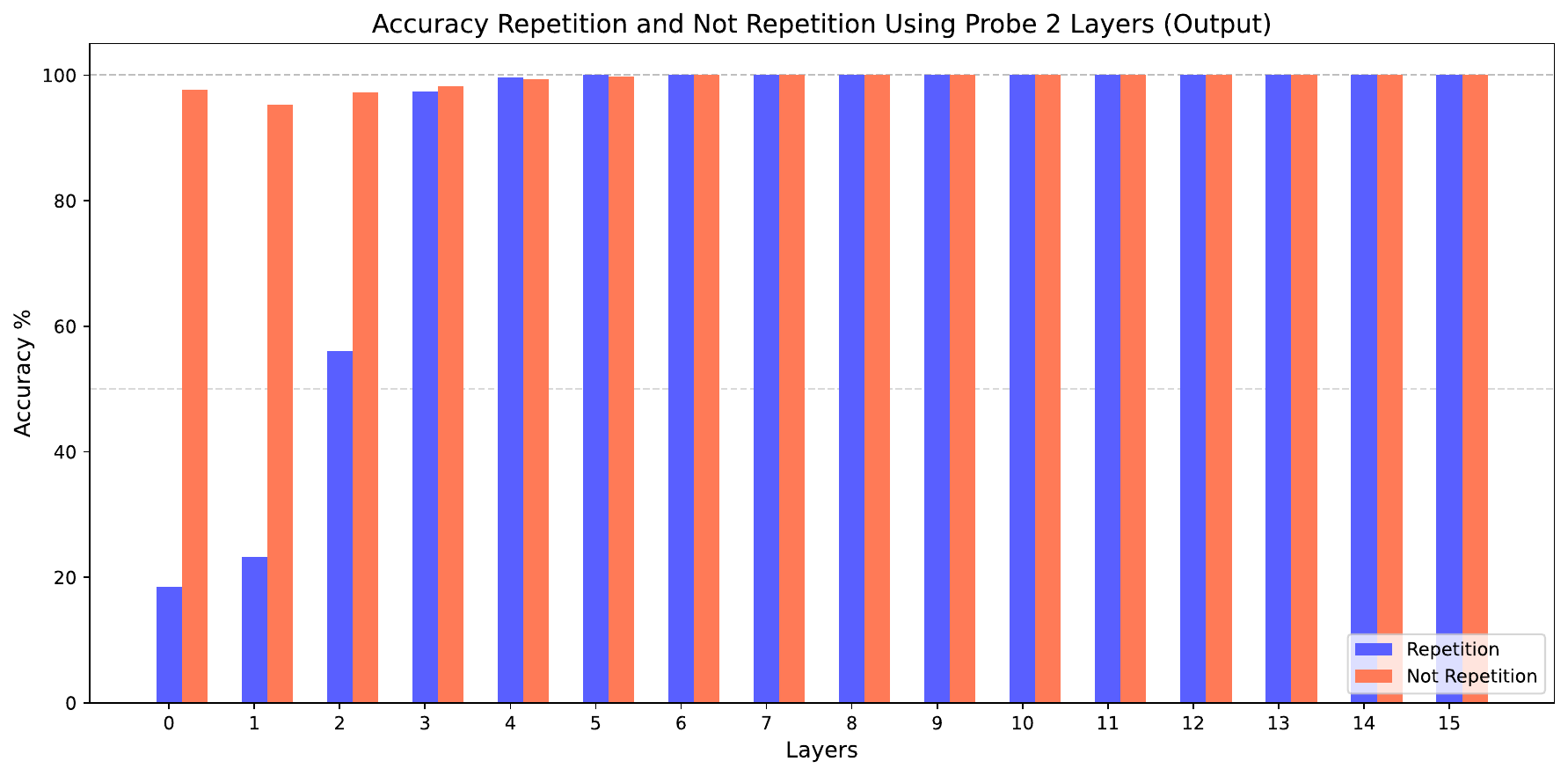}
    \caption{Classification Accuracy Using Two-Layer Probe on Output Activations for Repetition Detection}
    \label{fig:figure13}
\end{figure}

 The classification accuracy using the activations with the best Cohen's $d$ at each layer is presented in \autoref{fig:figure12}, while \autoref{fig:figure13} shows the accuracy achieved using our two-layer probe.

Similar to our findings with memorization, we achieved nearly 100\% accuracy both with individual activations that effectively separate the two groups and with our trained probes.

These findings demonstrate that our methodology extends beyond memorization detection. This suggests our approach could be valuable for studying other language model mechanisms like reasoning, knowledge retrieval, pattern matching, translation, physical world understanding, and more.

To further understand the mechanisms behind repetition detection, we analyzed the role of attention in this process. While our probe-based method effectively identifies repetition, examining the attention patterns reveals additional insights into how the model processes repeated sequences at different layers.

Attention heads play a crucial role in repetition, as they are responsible for identifying previous token repetitions. In earlier layers, many attention heads are dedicated to focusing exclusively on tokens identical to themselves. In later layers, this focus shifts to the next token in the repeated sequence. In some attention heads, the repeated token or its prediction dominates the attention values; in others, the mechanism becomes more complex, such as attending to the entire repeated sequence or its final tokens.

It is important to note that the attention heads responsible for identifying repetition are often not the same as those that perform better in repetition detection. This discrepancy occurs because attention heads that focus on repetition may lose much of the contextual information, concentrating more on the repetition mechanism itself. In contrast, the heads that are better at detection retain more of the context, which is where the detection features reside.

This rich contextual understanding enables our method to go beyond simple pattern matching. A significant advantage of using methods based on the model's internal representations over traditional approaches that search for verbatim repetition is the ability to detect repetition even when it is not identical word-for-word. This scenario occurs frequently with URLs. For example, a sample might include a blog post titled ``Top Destinations to Visit in Europe'' and within the text, there may be a URL like ``www.website.com/top-destinations-to-visit-in-europe''. A word-by-word method would not detect this repetition, but our probe can easily identify it.

\section{Evaluation}
\label{sec:evaluation}

During the evaluation of our probes, we identified sequences that were classified as memorized by the probe but exhibited an average cross-entropy loss greater than 2, as discussed in Section~\ref{sec:labeling_tokens}. To understand the reasons behind this discrepancy, we manually analyzed 1,000 of these sequences. The results of this analysis are summarized in \autoref{table:analysis}.

\begin{table}[h]
    \capstart
    \centering
    \begin{tabular}{l c}
        \hline
        \textbf{Category} & \textbf{Percentage (\%)} \\
        \hline
        Few large losses & 79.2 \\
        Calls to action & 10.6 \\
        Disclaimers & 7.4 \\
        Others & 2.8 \\
        \hline
    \end{tabular}
    \caption{Analysis of Sequences Classified as Memorized with High Cross-Entropy Loss}
    \label{table:analysis}
\end{table}

The \textbf{``Few large losses''} category comprises samples that are generally memorized but contain a small number of tokens with high loss values, which increases the overall average loss of the sequence. This situation often arises when the sequence follows a specific format that is reused frequently but includes variable elements. For instance, websites might use a standardized template for different entities, altering only the name and specifications automatically. Although these sequences were flagged as potential misclassifications due to our loss threshold, they are, in fact, memorized.

The \textbf{``Calls to action''} (e.g., ``Click here to\ldots'') and \textbf{``Disclaimers''} (e.g., ``By clicking next, you agree that we\ldots'') are short sequences that the probe classified as memorized but are not truly memorized in the traditional sense. We hypothesize that this misclassification occurs because these sequences exhibit highly similar patterns and structures, leading the model to incorrectly identify them as memorized.

The \textbf{``Others''} category includes a small number of sequences with unique patterns for which we could not ascertain why the model employs the memorization mechanism.

It is important to highlight that in all these cases, the activations that distinguish memorized sequences, as discussed in Section~\ref{sec:detecting_activations}, also classified them as memorized. Therefore, the issue does not lie with the probes themselves.

To mitigate this problem, we trained additional probes using the cross-entropy loss as labels. Instead of training the model to predict a binary label, memorized (1) or not memorized (0), we trained the probes to predict a continuous value inversely related to the loss of the token. Specifically, we assigned a label of 0 for not memorized tokens and used the loss value for memorized tokens, which typically ranges from 0 to 10. We clipped the loss at a maximum of 2 because beyond this point, the token is clearly not memorized. To accentuate the difference between memorized and not memorized tokens, we squared the result. The labeling formula is as follows:

\begin{equation}
\label{eq:label_formula}
\text{label} = \left(1 - \frac{\min(\text{loss}, 2)}{2} \right)^2
\end{equation}

With this formula, losses close to 0 are labeled close to 1 (memorized), while losses close to 2 are labeled close to 0 (not memorized).

The results demonstrated that the new probe could effectively differentiate between sequences that are fully memorized and those that utilize the memorization mechanism but fail to make accurate predictions. Specifically, the average probe value for misclassified tokens decreased from 0.91 to 0.27, while the average for correctly memorized tokens slightly decreased from 0.94 to 0.82. This indicates that the probe maintains its ability to detect memorized tokens while reducing misclassifications. It is important to note that we are reporting average values; for most memorized tokens, the probe still outputs a very high memorization score. The new probe is simply more cautious in its predictions and accounts for degrees of memorization.

As mentioned, the new probe also aids in detecting varying degrees of memorization. We observed that many sequences are not fully memorized (i.e., with a cross-entropy loss very close to 0) but are partially memorized. In these cases, the model predicts the next token with high probability but is not entirely confident. For example, in a fully memorized sequence, the model might predict the token ``provide'' with nearly 100\% probability in the context ``[\ldots] from the information which you \textit{provide}.'' In a slightly memorized sequence, the model might assign 80\% probability to ``provide,'' 10\% to ``supply,'' and distribute the remaining probability among similar tokens. Slightly memorized sequences are characterized by such distributions across most tokens, whereas fully memorized sequences have only a few tokens with lower confidence. The probe trained with the loss labels can effectively differentiate between these types.

Aside from these refinements, the probes have demonstrated exceptional performance in classifying memorization and repetition, as illustrated in the plots presented in Section~\ref{sec:results}.

To ensure that the probes are not overlooking edge cases, we examined all sequences with an average loss smaller than 1 that were not classified as memorized or repetition. We did not find any sequences that did not fall into either of these two categories or another known complementary mechanism. This reinforces the robustness of our probes in accurately detecting memorization within the model.

\subsubsection{Robustness}

We evaluated both the robustness of the model in maintaining memorization when faced with perturbed samples and the robustness of our probes in identifying such memorization. To this end, we applied various perturbations to the samples.

All tests were conducted using our dataset of speeches, focusing on the first 20 reliable tokens. These tokens have a suitable average length and are highly verbatim memorized.

\paragraph{Not Memorized Sequences}

First, we investigated whether we could induce memorization by attempting to deceive the model into believing that a not memorized sequence is memorized.

The perturbations applied are as follows:

\begin{itemize}
    \item Prepending ``Here is a sample''
    \item Prepending ``Here is a text''
    \item Prepending ``Here is a random text''
    \item Prepending ``Here is a very famous speech''
    \item Prepending ``Here is a text from a famous book''
    \item Prepending ``Here is a passage from the Bible''
    \item Inserting five memorized speeches above the target sequence
    \item Inserting five not memorized speeches above the target sequence
\end{itemize}

We then measured the values of the memorization probe at the output of each layer, as shown in \autoref{table:not_memorized}. The memorization values range from 0 (not memorized) to 100 (memorized). Since we are examining not memorized sequences, the values should ideally approach 0 at the end.

\begin{table}[h]
    \capstart
    \centering
    \begin{tabular}{lccccc}
        \hline
        \textbf{Perturbation} & \textbf{Layer 0} & \textbf{Layer 3} & \textbf{Layer 5} & \textbf{Layer 9} & \textbf{Layer 15} \\
        \hline
        Baseline & 53.9 & 32.3 & 14.3 & 0.2 & 0.0 \\
        Sample & 46.4 & 28.6 & 15.3 & 0.1 & 0.0 \\
        Text & 63.8 & 46.1 & 30.4 & 0.1 & 0.1 \\
        Random text & 56.2 & 33.3 & 15.5 & 0.1 & 0.1 \\
        Famous speech & 67.6 & 61.6 & 53.2 & 8.4 & 1.4 \\
        Famous book & 71.4 & 61.6 & 57.2 & 1.2 & 0.8 \\
        Bible & 85.9 & 65.7 & 58.4 & 8.7 & 2.2 \\
        5 memorized speeches & 87.5 & 71.3 & 76.4 & 52.0 & 25.1 \\
        5 not memorized speeches & 44.8 & 20.4 & 3.6 & 0.4 & 0.0 \\
        \hline
    \end{tabular}
    \caption{Memorization Probe Values for Not Memorized Sequences Under Various Perturbations}
    \label{table:not_memorized}
\end{table}

In the baseline case, the memorization values start around 50 (tie) and decrease to 0 across the layers, representing the model's normal behavior on not memorized sequences.

Our results demonstrate that it is indeed possible to create an illusion of memorization within the model's representations by manipulating the input. We observe that in the early layers, the effect is particularly strong when using tokens commonly associated with memorized samples, such as ``Bible,'' ``famous speech,'' and ``famous book.''

Conversely, adding the token ``random'' (e.g., ``Here is a random text'') reduces the memorization values in the early layers when compared to ``Here is a text,'' indicating that the model is less inclined to treat the sequence as memorized.

Notably, inserting five memorized speeches before the not memorized sequence substantially increases memorization values to 52 at Layer 9 and 25 at Layer 15. However, these high values only apply to the first 20 reliable tokens, after which they typically drop to 0 as the model recognizes the sequence is actually not memorized.

Another critical observation is that despite the variations in memorization values across different perturbations, the model's predictions remain largely unchanged, even in the case where five memorized speeches precede the not memorized sequence.

\paragraph{Memorized Sequences}

We conducted a similar analysis on memorized sequences, attempting to make the model perceive them as not memorized.

The perturbations applied are as follows:

\begin{itemize}
    \item Prepending ``Here is a random text''
    \item Prepending ``Here is a text from [incorrect source]''
    \item Inserting the memorized speech into the middle of a random book
    \item Inserting five not memorized speeches above the target sequence
    \item Inserting five versions of the speech rewritten with synonyms above it
    \item Inserting five versions of the speech rewritten with synonyms above it, with the target speech also rewritten
\end{itemize}

The results are presented in \autoref{table:memorized}.

\begin{table}[h]
    \capstart
    \centering
    \begin{tabular}{lcccccc}
        \hline
        \textbf{Perturbation} & \textbf{Layer 0} & \textbf{Layer 3} & \textbf{Layer 5} & \textbf{Layer 9} & \textbf{Layer 15} & \textbf{Loss} \\
        \hline
        Baseline Memorized & 64.2 & 79.9 & 82.6 & 99.8 & 98.6 & 0.15 \\
        Random text & 63.6 & 78.4 & 82.2 & 99.7 & 98.6 & 0.15 \\
        Wrong source & 65.2 & 81.2 & 82.9 & 99.8 & 98.9 & 0.14 \\
        Middle of book & 58.0 & 66.7 & 65.8 & 90.2 & 89.9 & 0.66 \\
        5 not memorized speeches & 66.5 & 73.5 & 80.3 & 99.8 & 99.5 & 0.17 \\
        5 synonymous speeches & 61.3 & 32.8 & 12.4 & 50.4 & 47.2 & 0.92 \\
        
        \makecell[tl]{5 synonymous speeches \\ (target also rewritten)} & 58.4 & 12.0 & 3.2 & 0.2 & 4.2 & 1.26 \\
        \hline
    \end{tabular}
    \caption{Memorization Probe Values and Loss for Memorized Sequences Under Various Perturbations}
    \label{table:memorized}
\end{table}

For reference, when applying the last perturbation to not memorized sequences, the average cross-entropy loss was 1.7. Comparing this to the 1.26 loss for the memorized sequences indicates that even when a memorized sequence is rewritten with synonyms, it retains some degree of memorization.

Our findings suggest that it is challenging to perturb the memorized sequences effectively. Simple methods, such as prepending ``Here is a random text'' or attributing the text to an incorrect source, do not significantly impact the memorization mechanism. It is worth noting that we are presenting aggregate metrics; in some specific instances, these perturbations can affect early layers, as some memorized sequences only exhibit strong memorization signals after a few layers, while others do so from the very beginning.

\begin{figure}[h]
    \capstart
    \centering
    \includegraphics[width=0.8\textwidth]{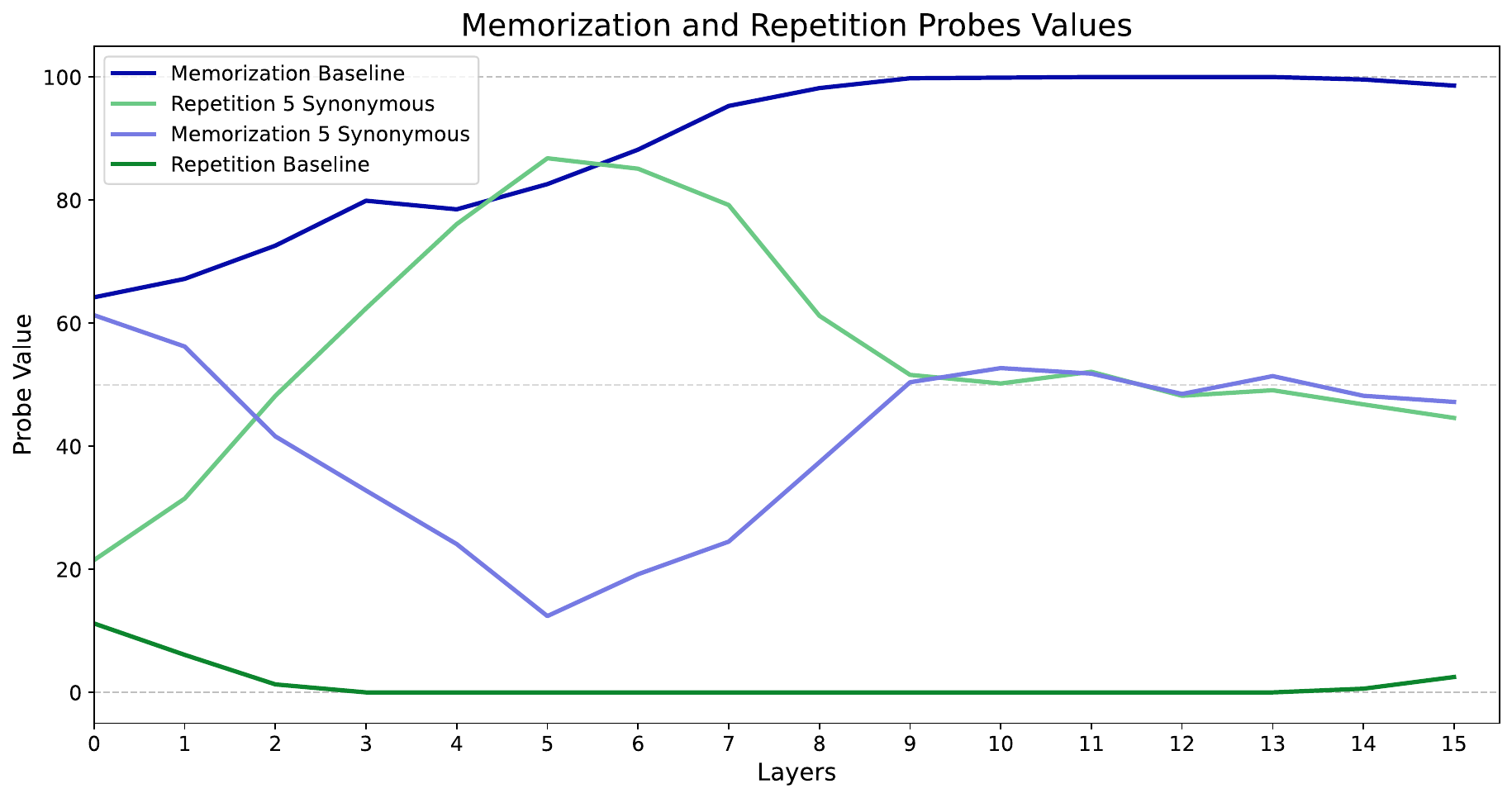}
    \caption{Memorization and Repetition Probe Values Under Synonymous Perturbation. The ``tug-of-war'' between the two mechanisms is very clear, and it is interesting to see that they really push one another, since when one mechanism goes up, the other goes down, and vice versa.}
    \label{fig:figure14}
\end{figure}

An unexpected result was that inserting synonymous versions of the speech above the target sequence substantially reduces the memorization values and increases the loss. As shown in \autoref{fig:figure14}, this effect likely occurs because the repetition mechanism overrides the memorization mechanism. We cannot be certain whether this represents a clear competition between the two mechanisms or another phenomenon; however, it is evident that the repetition mechanism plays a significant role.

As illustrated in \autoref{fig:figure14}, there appears to be a tug-of-war between the memorization and repetition mechanisms. In the baseline, the memorization mechanism strengthens across layers, while the repetition mechanism remains low. When we prepend synonymous copies of the sequence, the memorization values decrease, and the repetition values increase. Prepending exact copies amplifies the repetition mechanism significantly, causing the memorization values to diminish further.

From these observations, we infer that the model prefers to rely on repetition over memorization, which is sensible in many cases. This preference can lead to drawbacks, such as the increased loss observed in \autoref{table:memorized}, where the memorized sequence incurs a higher loss than the baseline. It appears that although the model is capable of predicting the memorized sequence, the presence of synonymous copies misleads it into being less confident in its memory and relying more on the repetition mechanism.

Additionally, when exact copies of the sequence are used, the memorization mechanism becomes even weaker, as seen in \autoref{table:memorization_repetition}. While one might argue that memorization still plays a role even when repetition is maximized, we cannot conclusively assert this because the initial copy involves memorization without repetition, and we know that memorization can be ``contagious,'' as seen when prepending five memorized sequences before a not memorized one.

\begin{table}[h]
    \capstart
    \centering
    \begin{tabular}{lccccc}
        \hline
        \textbf{Perturbation} & \textbf{Layer 0} & \textbf{Layer 3} & \textbf{Layer 5} & \textbf{Layer 9} & \textbf{Layer 15} \\
        \hline
        \multicolumn{6}{l}{\textbf{Memorization Probe Values}} \\
        Baseline & 64.2 & 79.9 & 82.6 & 99.8 & 98.6 \\
        5 synonymous & 61.3 & 32.8 & 12.4 & 50.4 & 47.2 \\
        5 exact copies & 60.4 & 9.6 & 15.2 & 37.6 & 40.8 \\
        \multicolumn{6}{l}{\textbf{Repetition Probe Values}} \\
        Baseline & 11.2 & 0.0 & 0.0 & 0.0 & 2.5 \\
        5 synonymous & 21.5 & 62.4 & 86.8 & 51.6 & 44.6 \\
        5 exact copies & 4.2 & 100.0 & 100.0 & 100.0 & 100.0 \\
        \hline
    \end{tabular}
    \caption{Memorization and Repetition Probe Values Under Perturbations}
    \label{table:memorization_repetition}
\end{table}

It is crucial to emphasize that not all sequences behave identically. In some cases, sequences with strong initial memorization values maintain higher memorization values even under perturbations.

\subsubsection{Additional Perturbation Tests}

We also conducted tests where we inserted a token between every word. This was done using periods (dots) and line breaks as controls. For example, the sentence:

\begin{quote}
\textbf{Original Sentence:}

``I have the best mom in the world''
\end{quote}

Becomes:

\begin{quote}
\textbf{Perturbed Sentence:}

``I . have . the . best . mom . in . the . world''
\end{quote}

\begin{figure}[h]
    \capstart
    \centering
    \includegraphics[width=0.8\textwidth]{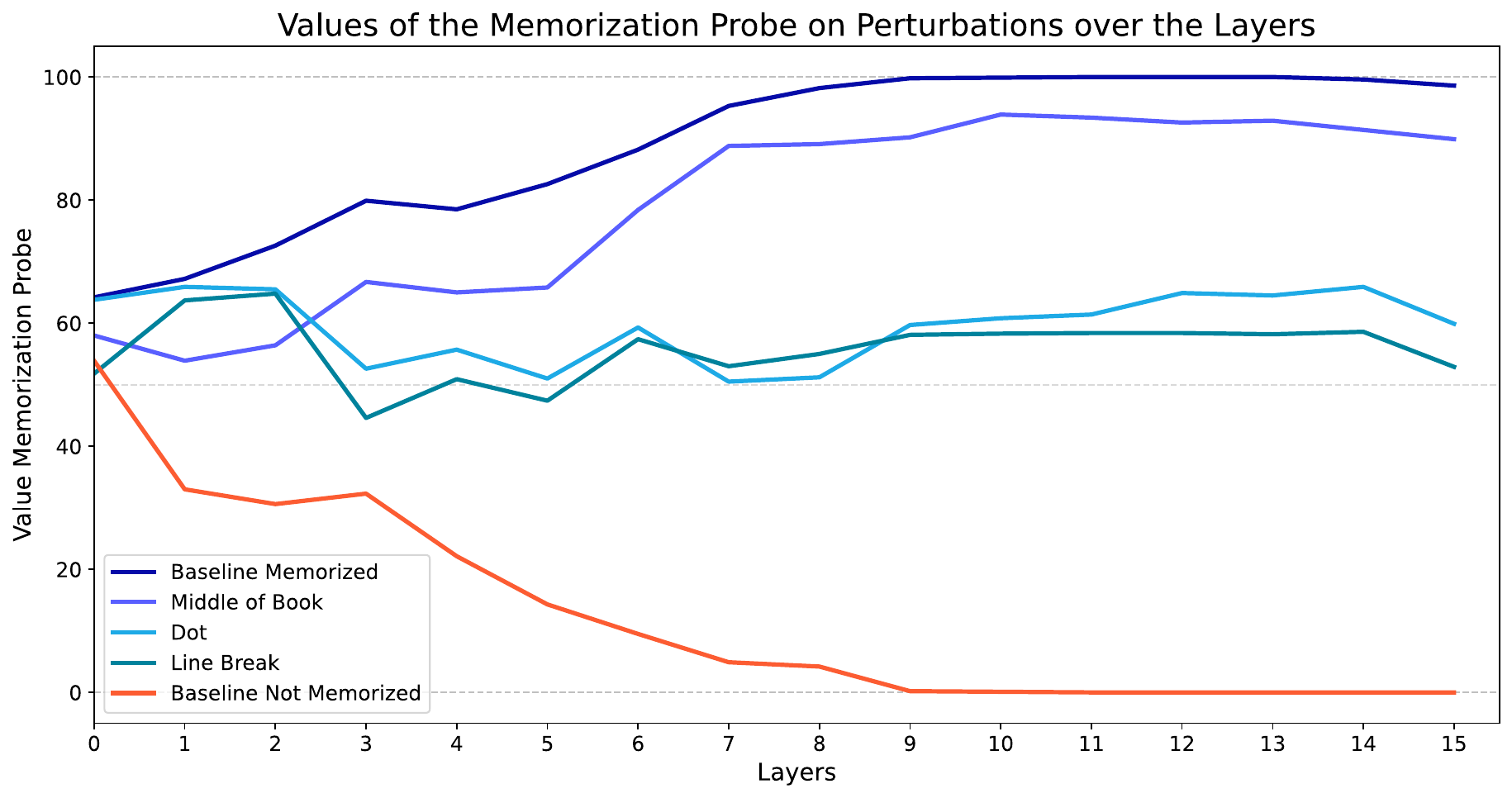}
    \caption{Memorization Probe Values Across Layers with Token Insertion Perturbations. Although the model still detects memorization, it is really affected when perturbed with tokens in between words.}
    \label{fig:figure15}
\end{figure}

\begin{figure}[h]
    \capstart
    \centering
    \includegraphics[width=0.8\textwidth]{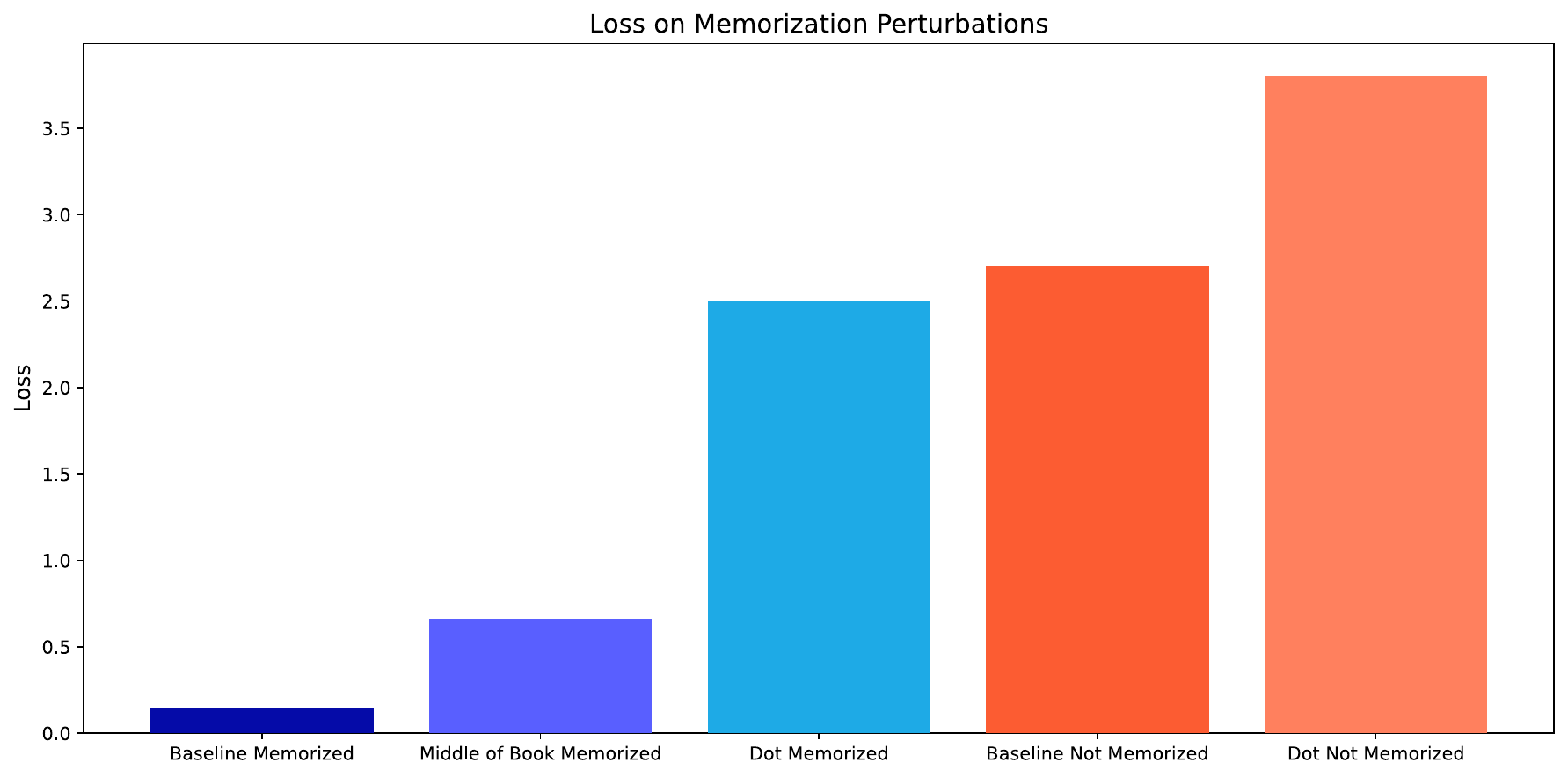}
    \caption{Cross-Entropy Loss with Memorization Perturbations. The ``Dot'' perturbation is really strong, and it affects the model's ability to predict the next token very heavily. However, this plot confirms that the model can still predict the next token better when memorized vs.\ not memorized, even with the ``Dot'' perturbations.}
    \label{fig:figure16}
\end{figure}

The results are depicted in Figures~\ref{fig:figure15} and \ref{fig:figure16}.

\paragraph{Observations}

Inserting characters between words disrupts the model's ability to accurately predict the next memorized token but does not significantly affect the identification of memorization, as seen in Figures~\ref{fig:figure15} and \ref{fig:figure16}. We hypothesize that this is because the next memorized token is typically retrieved based on the context provided by the previous token. For instance, in the sequence ``Twinkle, twinkle, little \ldots'', the prediction of ``star'' relies on the context of ``little.'' By inserting an additional token like a period, the model struggles to transfer this context effectively.

Supporting this, we also tested inserting a period every $n$ words. With a period inserted every ten words, the model could predict subsequent words with nearly 100\% accuracy, except for those immediately following the inserted token.

In these cases, other mechanisms seem to take over from memorization. For example, in the memorized phrase ``[\ldots] tears and \emph{blood},'' the model predicts ``blood'' with 100\% accuracy. When a period is inserted between ``and'' and ``blood,'' the prediction shifts to a 50\% probability for ``blood'' and 50\% for ``sweat.'' This suggests that the model struggles to maintain memorization across inserted tokens, sometimes predicting the memorized token accurately, sometimes partially, and sometimes not at all.

\section{Intervening}
\label{sec:intervening}

We leverage our trained probes to intervene in the model's activations to alter its behavior. Specifically, we demonstrate that we can suppress the memorization and repetition mechanisms, compelling the model to utilize alternative internal mechanisms for next-token prediction.

To attenuate memorization, we subtract the intervention from the activation set during the forward pass. The intervention is computed by projecting the activation vector onto the direction of the normalized probe weights, scaling the projection by a hyperparameter $\alpha$, and reconstructing the vector in that direction to create the final intervention. To better preserve not memorized tokens while effectively targeting memorized ones, we square the projection. We then subtract this computed intervention from the original activations. Mathematically, the intervention is defined as:

\begin{equation}
\label{eq:intervention}
\text{result} = \text{activations} - \alpha \left( \text{activations} \cdot \frac{W_{\text{probe}}}{\| W_{\text{probe}} \|} \right)^2  \frac{W_{\text{probe}}}{\| W_{\text{probe}} \|}
\end{equation}

The hyperparameter $\alpha$ can vary by activation type and layer. For instance, we may require a larger $\alpha$ for earlier layers and smaller values for later layers, as well as different magnitudes for the output layer compared to the dense attention computations. As observed by \citet{maini2023can}, memorization is distributed across many layers. Given the vast continuous search space this introduces, we employed a custom genetic algorithm. The optimization objective was to elevate the loss of memorized sequences to match that of not memorized ones, while keeping the loss of not memorized sequences unchanged.

\begin{figure}[H]
    \capstart
    \centering
    \includegraphics[width=0.8\textwidth]{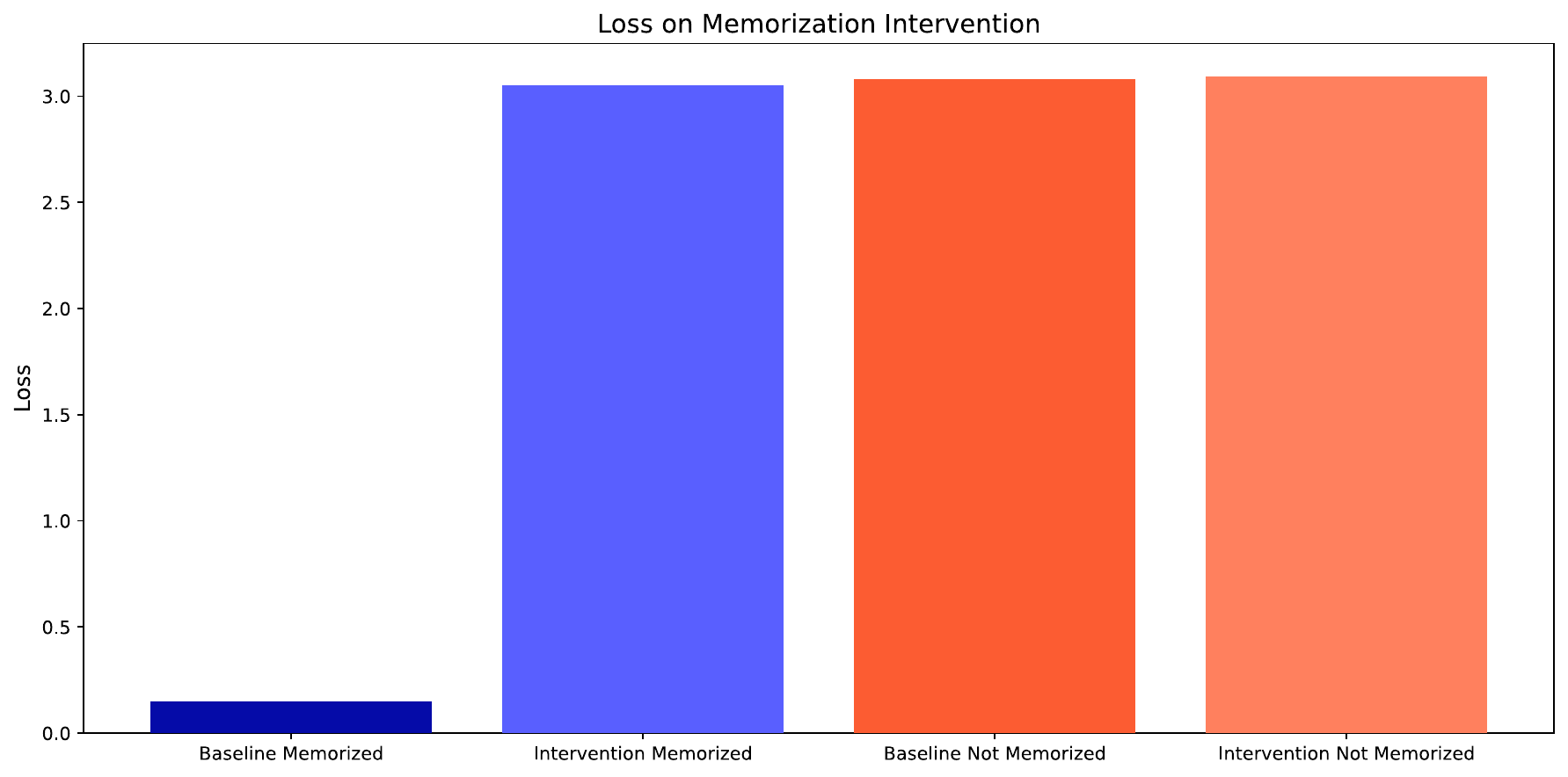}
    \caption{Loss on Memorization Intervention. By intervening in the mechanism, we were able to completely disable memorization while keeping all other mechanisms intact.}
    \label{fig:figure17}
\end{figure}

We then compared the sequences from our curated dataset of same-distribution samples, both memorized and not memorized, before and after intervention. The results are presented in \autoref{fig:figure17}.

As illustrated in \autoref{fig:figure17}, we effectively eliminated the memorization mechanism from the memorized samples, while the not memorized samples remained virtually unaffected.

\begin{figure}[h]
    \capstart
    \centering
    \includegraphics[width=0.8\textwidth]{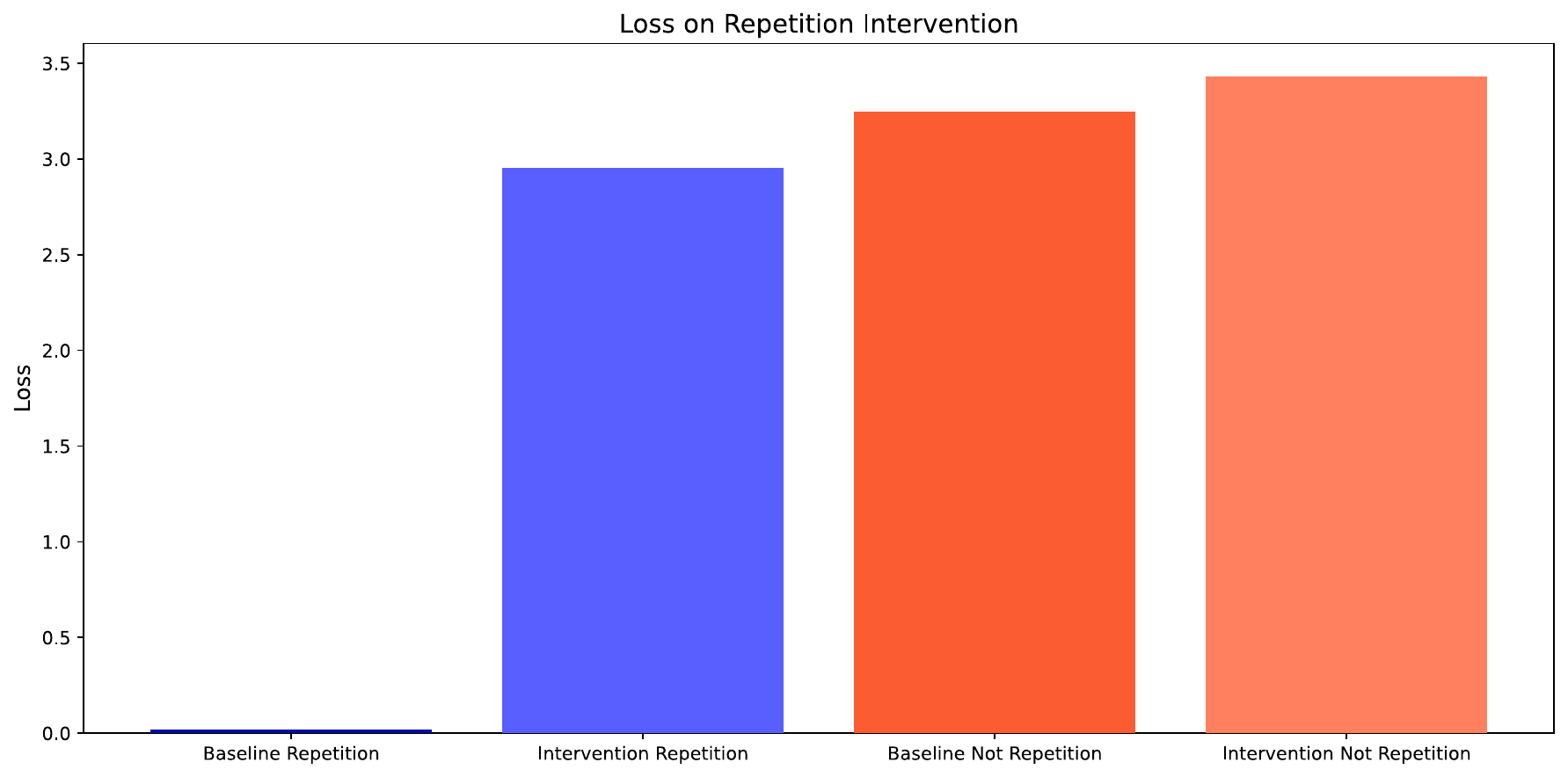}
    \caption{Loss on Repetition Intervention. The repetition mechanism is more robust to changes and needed stronger intervention. The intervention worked extremely well but was not enough to remove the mechanism completely. The increase in ``Intervention Not Repetition'' happened because there are actually some tokens that use repetition in normal general text.}
    \label{fig:figure18}
\end{figure}

We applied the same approach to the repetition mechanism, and the results are shown in \autoref{fig:figure18}.

Repetition appears to be a more robust mechanism within the model, making it harder to fully eliminate. Nevertheless, we were able to substantially reduce its influence by increasing the loss from nearly zero to about three, while maintaining the loss of samples without repetition at their original values. The increase in loss for "Intervention Not Repetition" can be attributed to common repetitions in general text, such as names and frequent expressions, which contributed to the observed increase.

In both cases, we ensured that the interventions preserved the model's coherence and overall ability to predict tokens, while specifically disabling the targeted mechanisms. For example, consider Martin Luther King Jr.'s famous ``I Have a Dream'' speech. In the following excerpt:

\begin{quote}
I have a dream that my four little children will one day live in a nation where they will not be judged by the color of their skin\ldots
\end{quote}

Under normal conditions, the model predicts all tokens with nearly 100\% accuracy. However, after intervention, instead of predicting ``\ldots\ one day live in a \emph{nation},'' the top five outputs become ``house'' (22\%), ``place'' (11\%), ``world'' (6\%), ``home'' (5\%), and ``city'' (4\%). This indicates that the model lost its ability to rely on memorization but maintained a highly coherent ability to predict using alternative mechanisms.

Notably, the effectiveness of our interventions reveals important insights about the underlying mechanisms. Although we trained the probes solely to detect memorization and repetition, they also proved capable of directly intervening in these mechanisms, suggesting the identified features have a causal impact on the model's behavior. This finding is particularly noteworthy given prior challenges in establishing such relationships. As \citet{durmus2024steering} caution in their work on feature steering, ``there is a disconnect between feature activation context and resulting behavior'', noting that feature detection may not directly correspond to feature intervention.

To illustrate our contrasting result: when determining if a car drove on a dirt road, one might look for wheel tracks, their presence suggests the car was there, while their absence suggests it wasn't. However, removing tracks or using wheels that don't leave marks doesn't change whether the car actually drove there. In this analogy, the classification relies on an artifact (tracks) rather than a causal feature (the driving). Our probes, in contrast, identify features that both correlate with and causally influence the mechanisms, demonstrating a more direct relationship between feature detection and behavioral control.

\section{Certainty}
\label{sec:certainty}

During our research, we discovered that the model has a robust mechanism for encoding certainty, reflected in the unique, high-magnitude activation of neuron 1668 in the residual stream from layer 4 onward. Tokens where activation 1668 has smaller values than its peers exhibit greater certainty about the next token. This includes mechanisms like memorization, repetition, completion, knowledge retrieval, and other categories where the model demonstrates certainty. Interestingly, we have found that the level of certainty can sometimes reveal the model's knowledge about specific content.

\begin{figure}[h]
    \capstart
    \centering
    \includegraphics[width=0.8\textwidth]{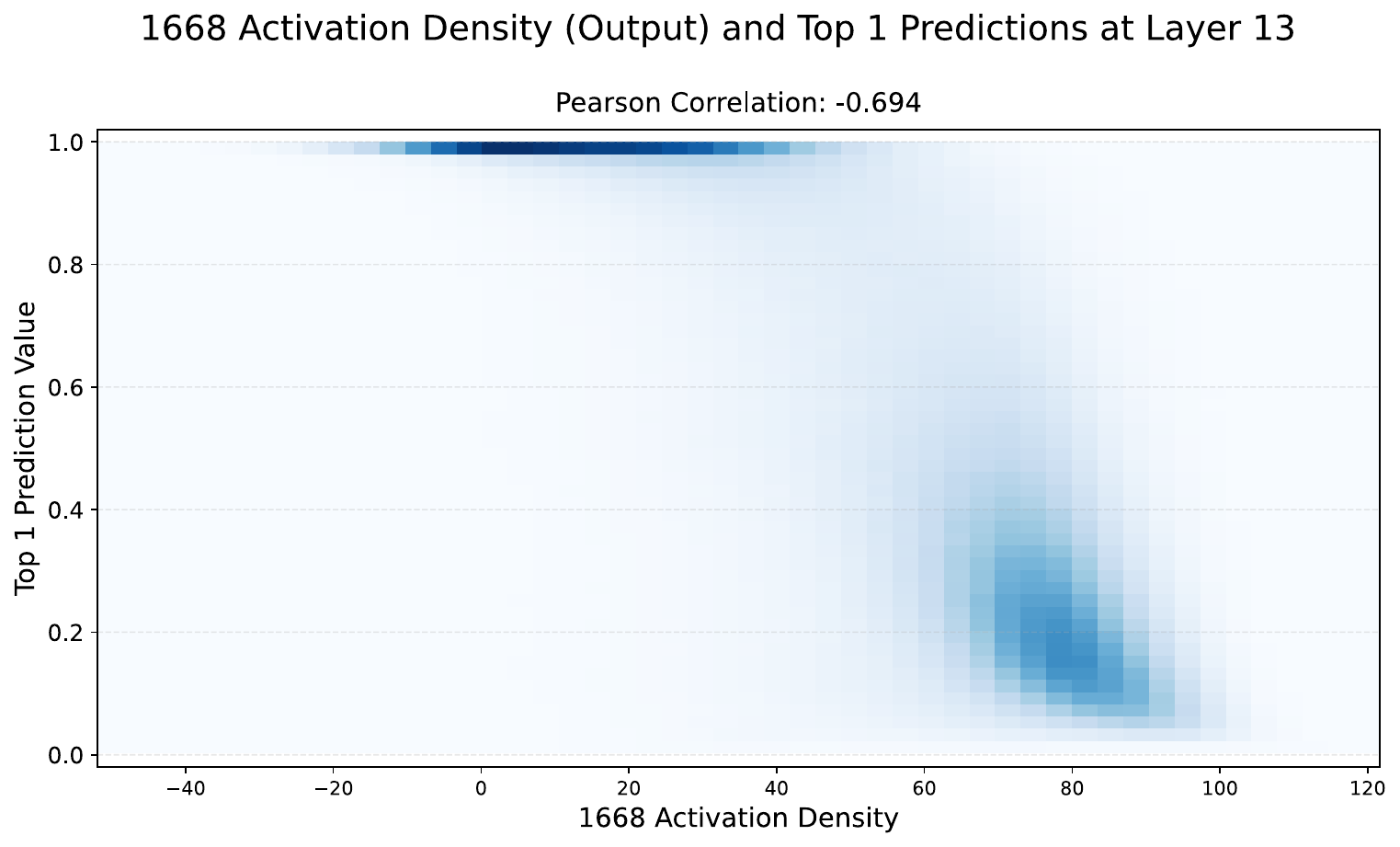}
    \caption{Density plot of activation 1668 values and top-1 prediction probabilities at layer~13 output. We observe a clear cluster of values with very high top-1 prediction probabilities when the activation 1668 value goes below 40; above this threshold, the probabilities decrease. This is confirmed by the strong negative Pearson correlation of almost $-0.7$.}
    \label{fig:figure19}
\end{figure}

To illustrate this, we present a plot comparing the values of activation 1668 at the output layer with the softmax probabilities of the top-1 predictions (see \autoref{fig:figure19}). There is a strong negative Pearson correlation of approximately $-0.7$ at layer~13, indicating that lower values of activation 1668 correspond to higher probabilities (closer to 100\%) of the top-1 prediction. This reflects a strong correlation between the activation value and the model's certainty in its predictions.

\begin{figure}[h]
    \capstart
    \centering
    \includegraphics[width=0.8\textwidth]{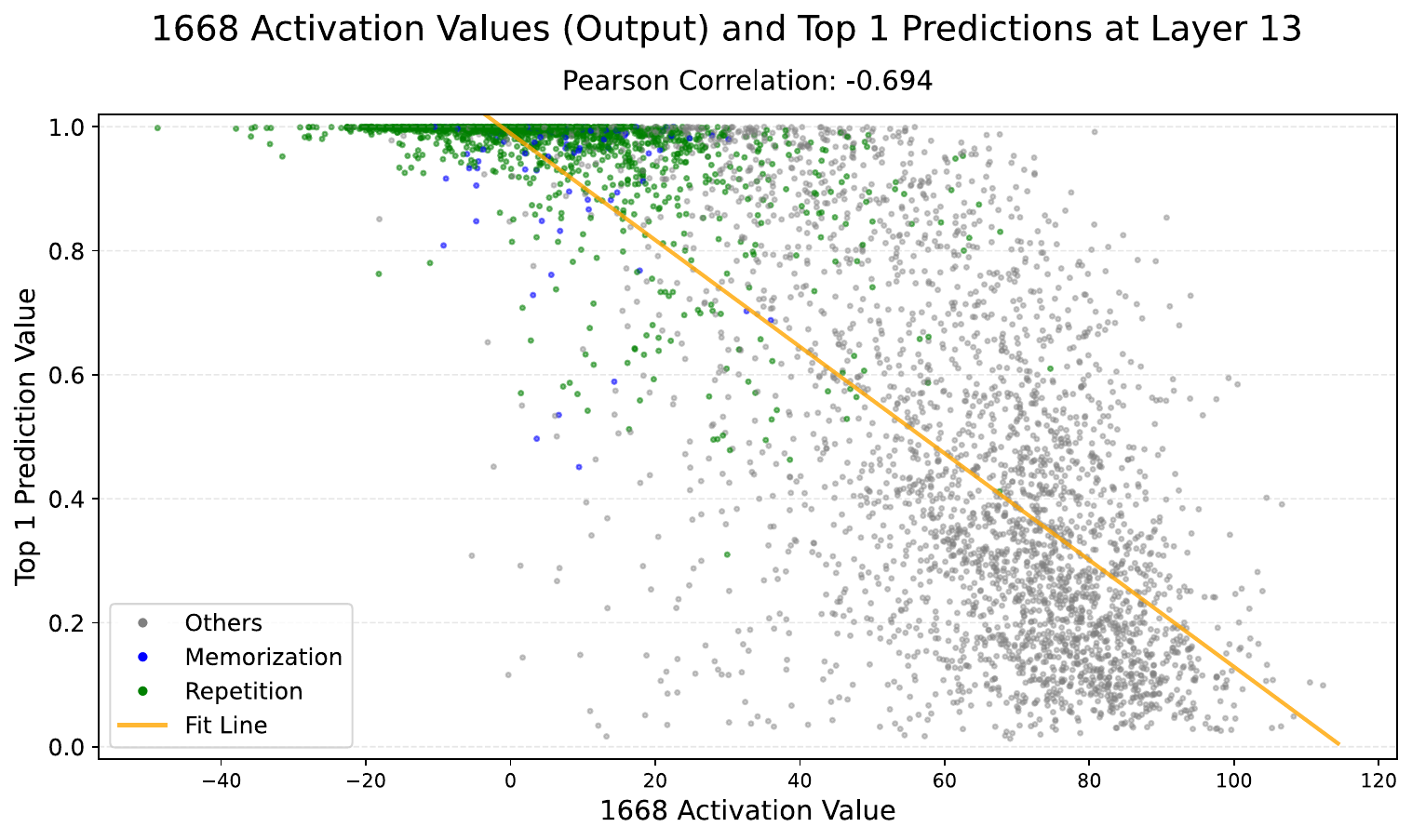}
    \caption{Activation 1668 values and top-1 prediction probabilities at layer~13 output. We see that almost all Repetition and Memorization tokens are concentrated in the high end of Top-1 Prediction and low end of activation 1668 value, showing that these mechanisms are highly correlated with the model's certainty on the prediction.}
    \label{fig:figure20}
\end{figure}

Furthermore, we observe that memorization and repetition mechanisms exhibit distinct distributions of activation 1668 values, indicating a high degree of certainty (see \autoref{fig:figure20}). This suggests that activation 1668 effectively differentiates between tokens involved in memorization, repetition, and other mechanisms that show certainty.

These insights into activation 1668's role in encoding certainty open up promising avenues for further research. By leveraging this activation pattern as a reliable indicator of model confidence, we can potentially develop more sophisticated methods for identifying and categorizing other mechanisms within the model. This could enable automated labeling of tokens based on certainty levels, leading to more nuanced inference strategies that dynamically adjust their approach based on the model's confidence. Such uncertainty-aware inference methods could be particularly valuable in applications requiring high reliability or in scenarios where graceful handling of uncertainty is crucial.

\begin{figure}[h]
    \capstart
    \centering
    \includegraphics[width=0.8\textwidth]{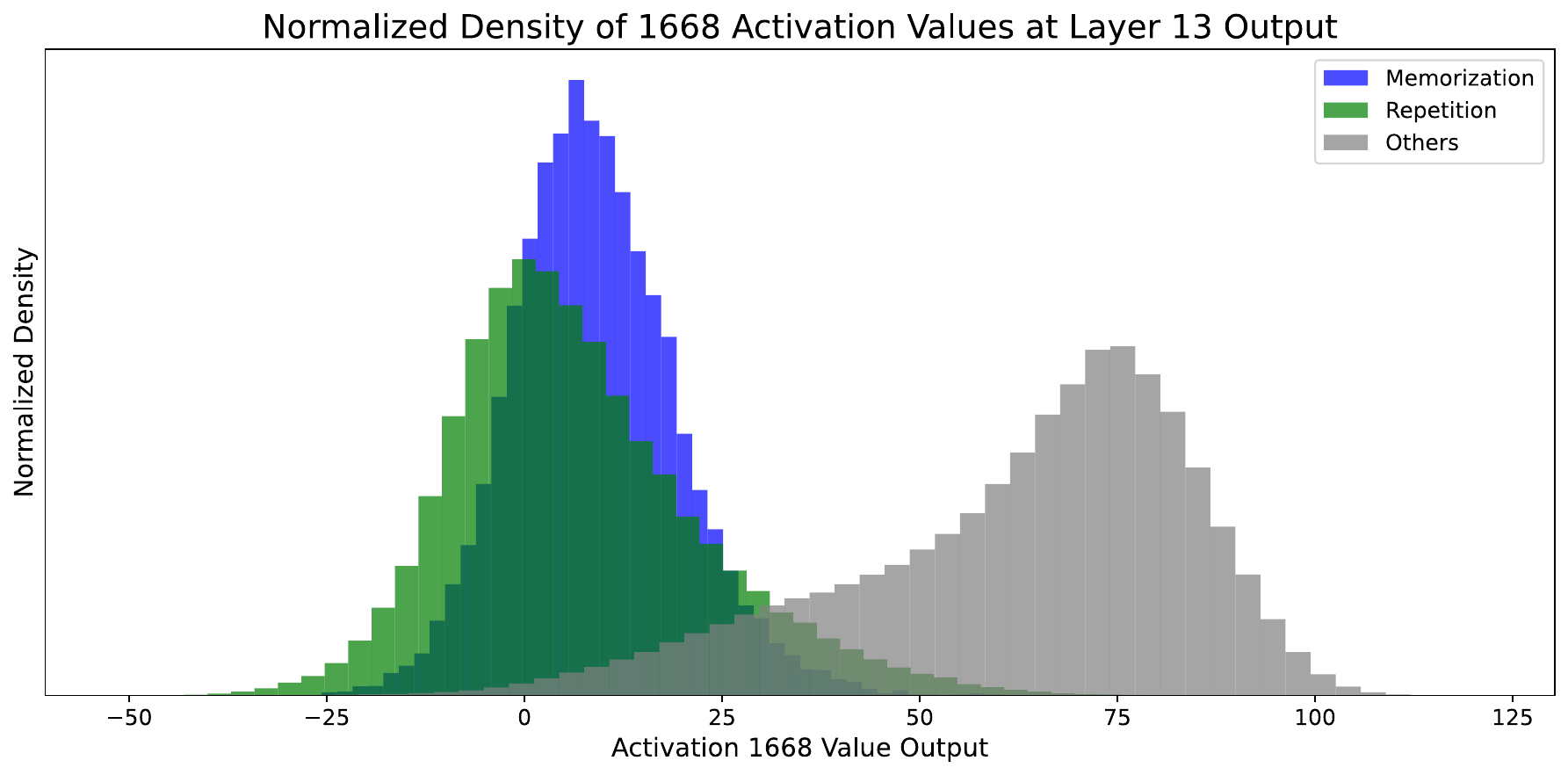}
    \caption{Normalized density of activation 1668 values for different mechanisms at layer~13 output. Confirming what \autoref{fig:figure20} shows, Memorization and Repetition have smaller values of the neuron 1668, which highlights greater certainty of the model.}
    \label{fig:figure21}
\end{figure}

In \autoref{fig:figure21}, we present the normalized density of activation 1668 values for memorization, repetition, and other tokens at the output of layer~13. The distributions confirm that tokens associated with memorization and repetition tend to have lower activation 1668 values, aligning with higher certainty in predictions.

\begin{figure}[h]
    \capstart
    \centering
    \includegraphics[width=0.8\textwidth]{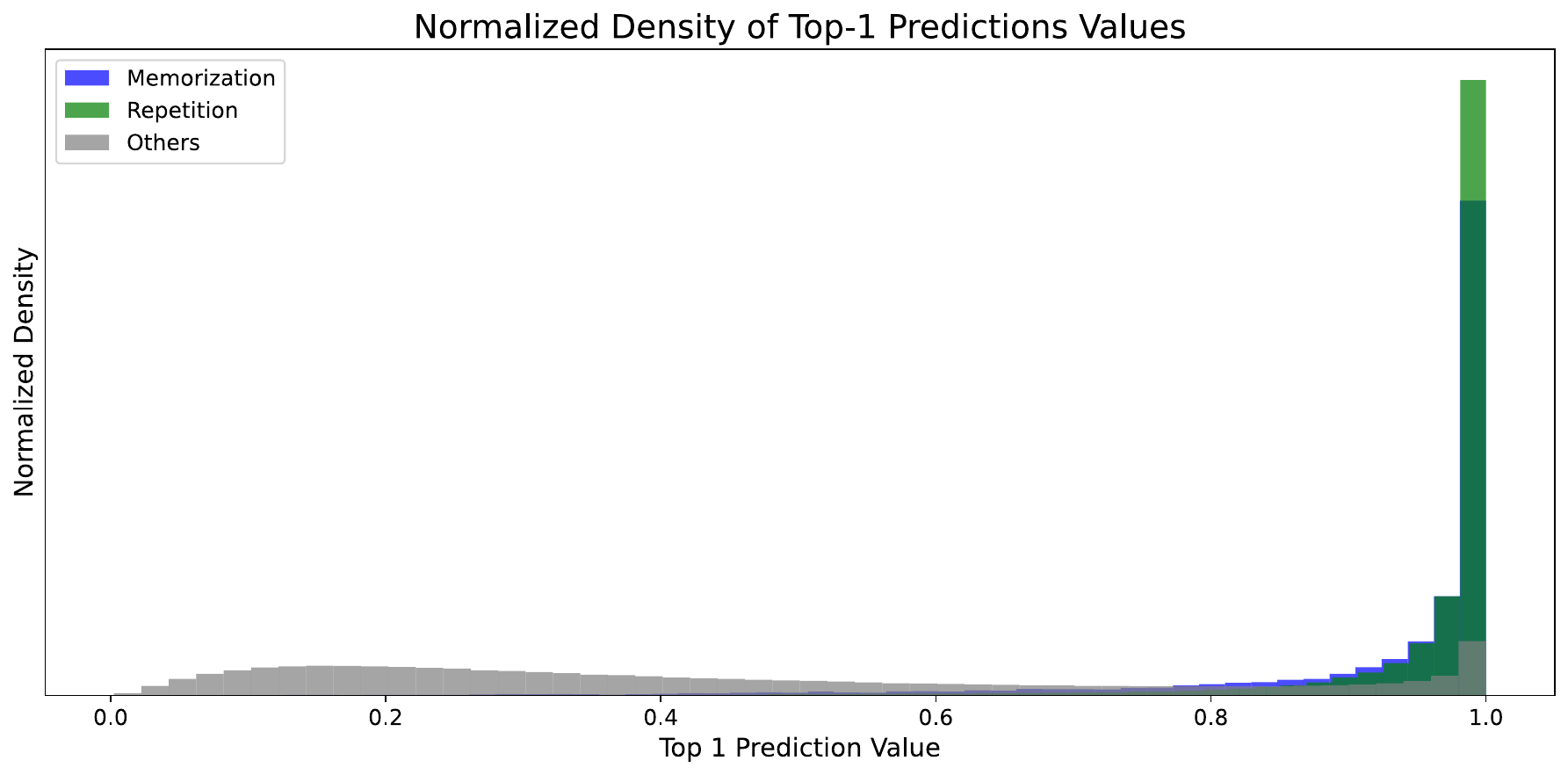}
    \caption{Normalized density of top-1 prediction probabilities for different mechanisms. Memorization and Repetition have a very high density of predictions close to 100\%.}
    \label{fig:figure22}
\end{figure}

Similarly, the distribution of top-1 prediction probabilities, shown in \autoref{fig:figure22}, reveals that tokens involved in memorization and repetition are more concentrated near 100\% probability compared to other tokens. This further supports the role of activation 1668 in representing the model's certainty.

\section{Interpretability}
\label{sec:interpretability}

As observed in \autoref{fig:figure11} in the Repetition section, more than 50\% of the activations at the output of Layer~11 effectively distinguish between tokens associated with repetition and those that are not. We consider it unlikely that such a large proportion of the token representation is dedicated solely to handling repetition. It is crucial to acknowledge that many activations are not exclusively utilized for the specific features we are measuring. For example, activation 1668 represents certainty and differentiates between repetition, memorization, and other mechanisms.

Even when accounting for this overlap, the fact that over half of the activations represent these kinds of mechanisms is significant. This behavior suggests the diverse ways in which large language models (LLMs) employ neurons as features. In this context, it is more plausible that the model uses these features as relative values rather than absolute ones. We observed this phenomenon in specific samples during our research, but were unable to find an aggregate measure.

Eventually, we identified a particularly interesting phenomenon in the intermediate activations of the multilayer perceptron (MLP). These activations not only separate memorization but also exhibit another distinct distribution exclusively for the token ``the,'' which, intriguingly, also separates memorization. This pattern occurs in several other activations for different tokens, such as ``an,'' ``to,'' ``for,'' etc.

\begin{figure}[h]
    \capstart
    \centering
    \includegraphics[width=0.8\textwidth]{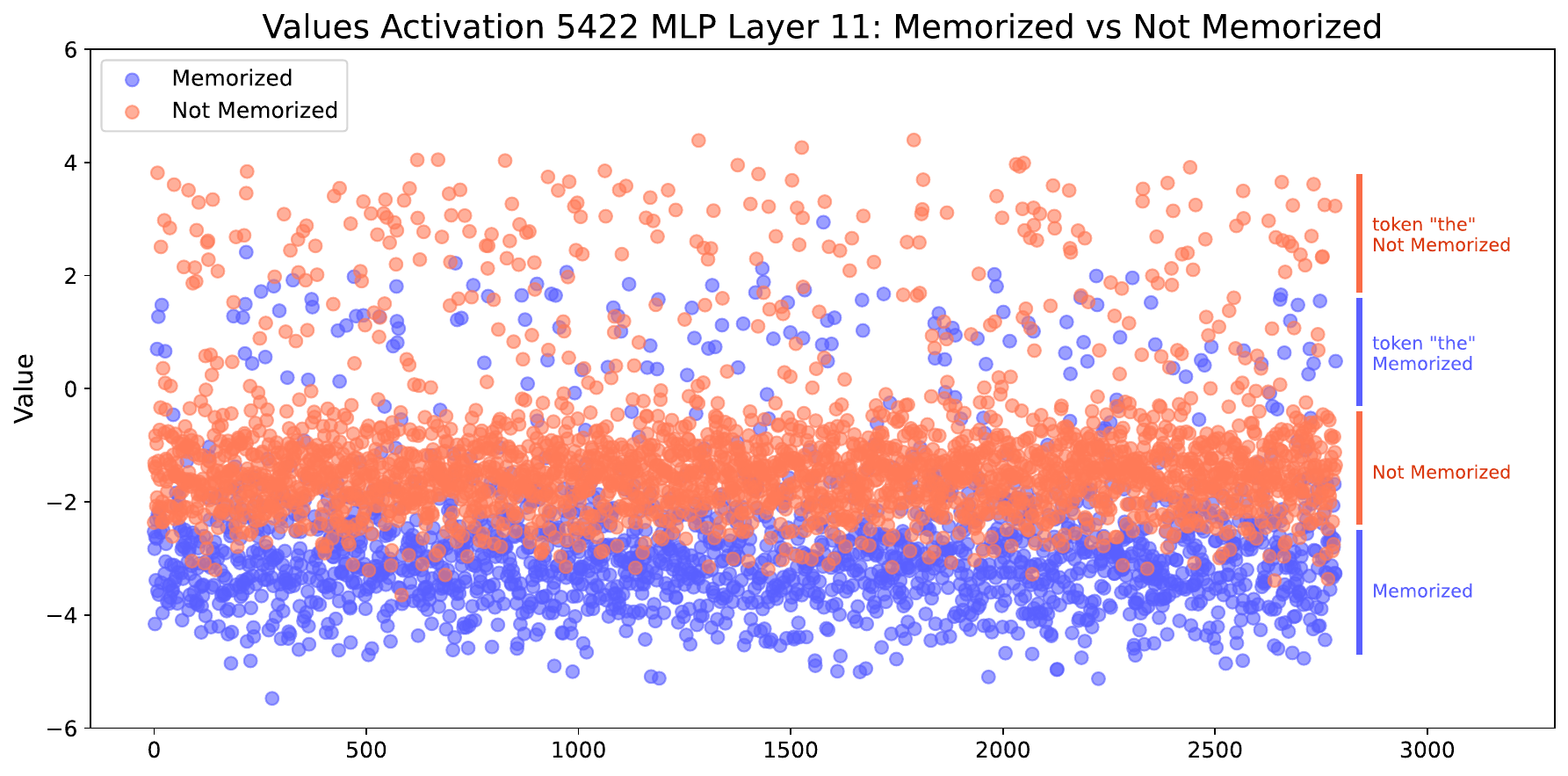}
    \caption{Scatter plot of activation values for neuron 5422 in MLP Layer~11. The plot compares memorized tokens, not memorized tokens, and occurrences of the token ``the.'' We observe that the activation values for ``the'' form a distinct cluster, separate from those of other memorized and not memorized tokens. This is one of the most interesting findings of our research. The model displays very different activation patterns for the token ``the'' (and also for other tokens such as ``an,'' ``to,'' etc.), which shows the complex way in which the model transforms the latent space and uses neuron activations as features.}
    \label{fig:figure23}
\end{figure}

\autoref{fig:figure23} illustrates the distribution of activation values for neuron 5422 in MLP Layer~11, comparing memorized and not memorized tokens. The figure shows that the activation values for the token ``the'' form a distinct distribution, separate from both memorized and not memorized tokens. This indicates that certain neurons are sensitive to specific tokens, in addition to their role in mechanisms like memorization.

This observation underscores the complexity of the internal representations within LLMs and highlights the importance of interpretability in understanding the functionality of these models. It suggests that the model may be utilizing these activations in a relative manner, adjusting their values based on context and token properties, rather than relying solely on absolute activation levels.

Moreover, the occurrence of similar patterns for other common tokens such as ``an,'' ``to,'' and ``for'' indicates that this is not an isolated phenomenon but may reflect a general characteristic of how the model encodes and differentiates between different linguistic elements and mechanisms like memorization.

\section{Discussion}
\label{sec:discussion}

We believe that the methodology presented in this paper---beginning with a small, diverse dataset of samples that can be distinguished by their activations, and then using these samples to label a larger, general dataset for training classifiers---can be effectively applied to other mechanisms within large language models (LLMs). While we have demonstrated its applicability to repetition, we anticipate that each mechanism will introduce its own nuances and challenges that must be addressed.

Perhaps the most impactful aspect of our approach is the ability to classify tokens based on the model's internal representations. This capability is extremely useful when selecting data for training models. For example, we can attenuate memorization mechanisms in mathematical problems, thereby encouraging the model to rely more on its mathematical reasoning processes.

We strongly advocate for more specialized training of LLMs to align them with specific use cases rather than solely focusing on downstream tasks. This approach is exemplified by practices such as fine-tuning or reinforcement learning from human feedback (RLHF; \citealp{christiano2017deep, stiennon2020learning, ouyang2022training}) applied to pre-trained models. Instead of merely predicting the next token in a general dataset, the model is trained to be more useful for its intended applications. Techniques that enable token labeling can be particularly powerful in achieving this objective.

We expect that the methods demonstrated in this paper will be effective for other types of mechanisms, such as factual retrieval, logical reasoning, mathematics, and so on. This is especially true when combined with other mechanisms like certainty, which can be traced back to the fundamental instances where they arise at the token level.

\section{Limitations}
\label{sec:limitations}

While our method achieves high accuracy in detecting memorization within LLMs, several limitations should be acknowledged. First, the probes we trained are specific to the model architecture and dataset used in our experiments. Although we anticipate that the underlying principles are applicable to other models, the probes may require adjustments when applied to different architectures or datasets.

Second, our method is primarily effective in detecting verbatim memorization. Identifying more nuanced forms of memorization, such as format-based or knowledge memorization, may be more challenging and require additional refinement of our techniques.

Lastly, despite extensive evaluations and tests, we recognize that the mechanisms identified by our probes might encompass more than just memorization or repetition. It is possible that these probes are capturing additional internal processes within the model, and further research is needed to fully disentangle and understand these underlying mechanisms.

\section{Conclusion}
\label{sec:conclusion}

In this paper, we introduced an analytical method for detecting memorization in large language models by focusing on their internal neuron activations. By identifying specific activations that effectively distinguish between memorized and not memorized tokens, we trained classification probes that achieved near-perfect accuracy in detecting memorization. Our approach not only provides a precise detection mechanism but also enhances interpretability by revealing how memorization manifests within the model's architecture.

We extended our methodology to detect other mechanisms, such as repetition, demonstrating the versatility of our approach in probing various internal processes of language models. Furthermore, we showed that it is possible to intervene in the model's activations to suppress specific mechanisms such as memorization and repetition, effectively altering the model's behavior without compromising its overall performance.

Our findings have significant implications for the development and evaluation of large language models. By providing tools to detect and control memorization, we enable better management of model behavior, ensuring that performance metrics genuinely reflect a model's capacity to generalize rather than its ability to recall training data. Additionally, the identification of a certainty mechanism within the model's activations opens avenues for further research into understanding and interpreting the internal states of language models.

Overall, our work contributes to the broader goal of improving model interpretability and reliability, offering practical methods for analyzing and intervening in the internal mechanisms of large language models.

\bibliographystyle{plainnat}
\bibliography{references}

\appendix
\section*{Appendix}

\section{Cohen's d Distribution Plots for Memorization}
\label{app:violin-plots-memorization}

\setlength{\intextsep}{5pt}
\setlength{\floatsep}{2pt}

\begin{figure}[!htb]
    \centering
    \includegraphics[width=\textwidth]{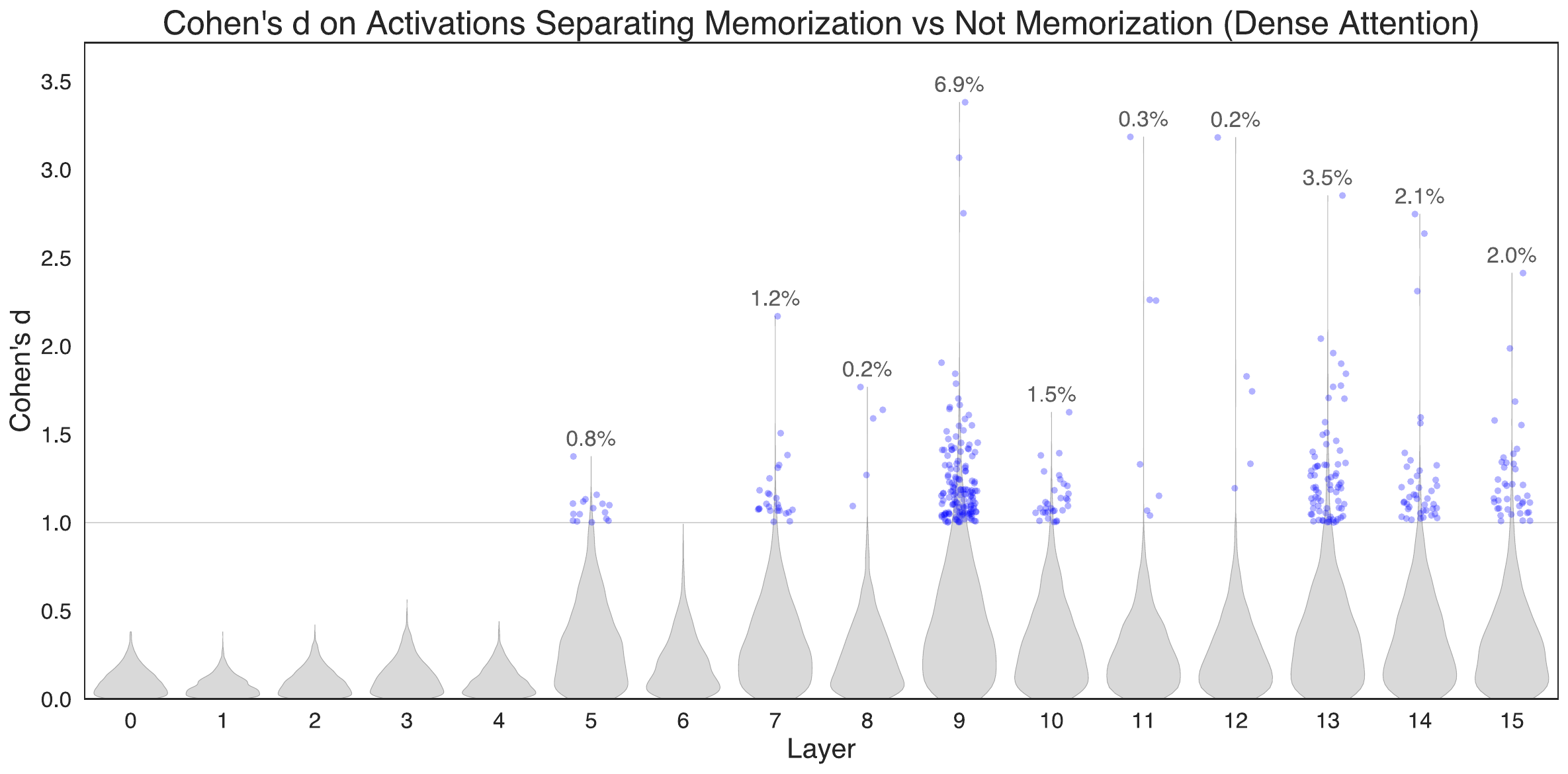}
\end{figure}

\begin{figure}[!htb]
    \centering
    \includegraphics[width=\textwidth]{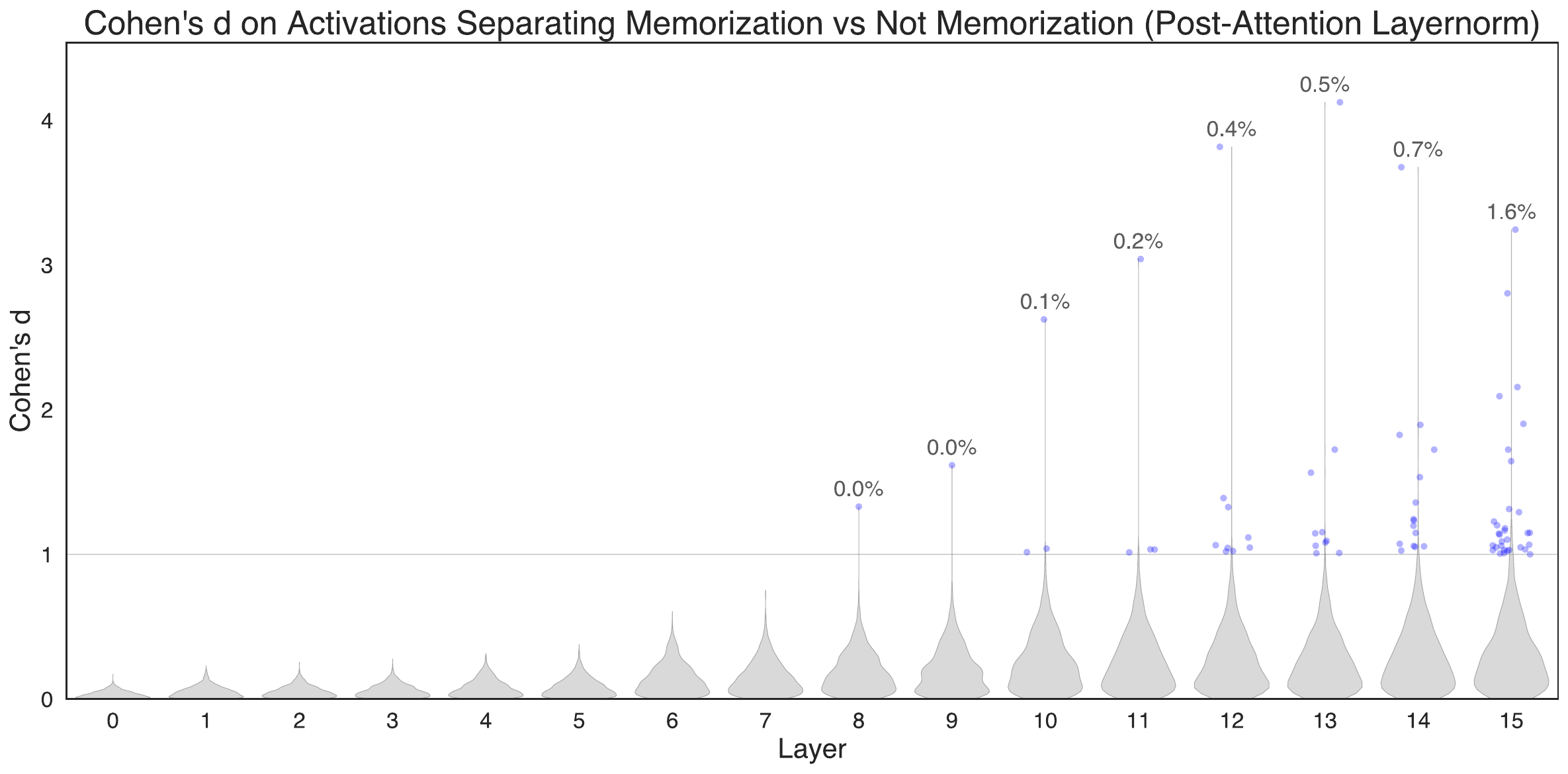}
\end{figure}

\begin{figure}[!htb]
    \centering
    \includegraphics[width=\textwidth]{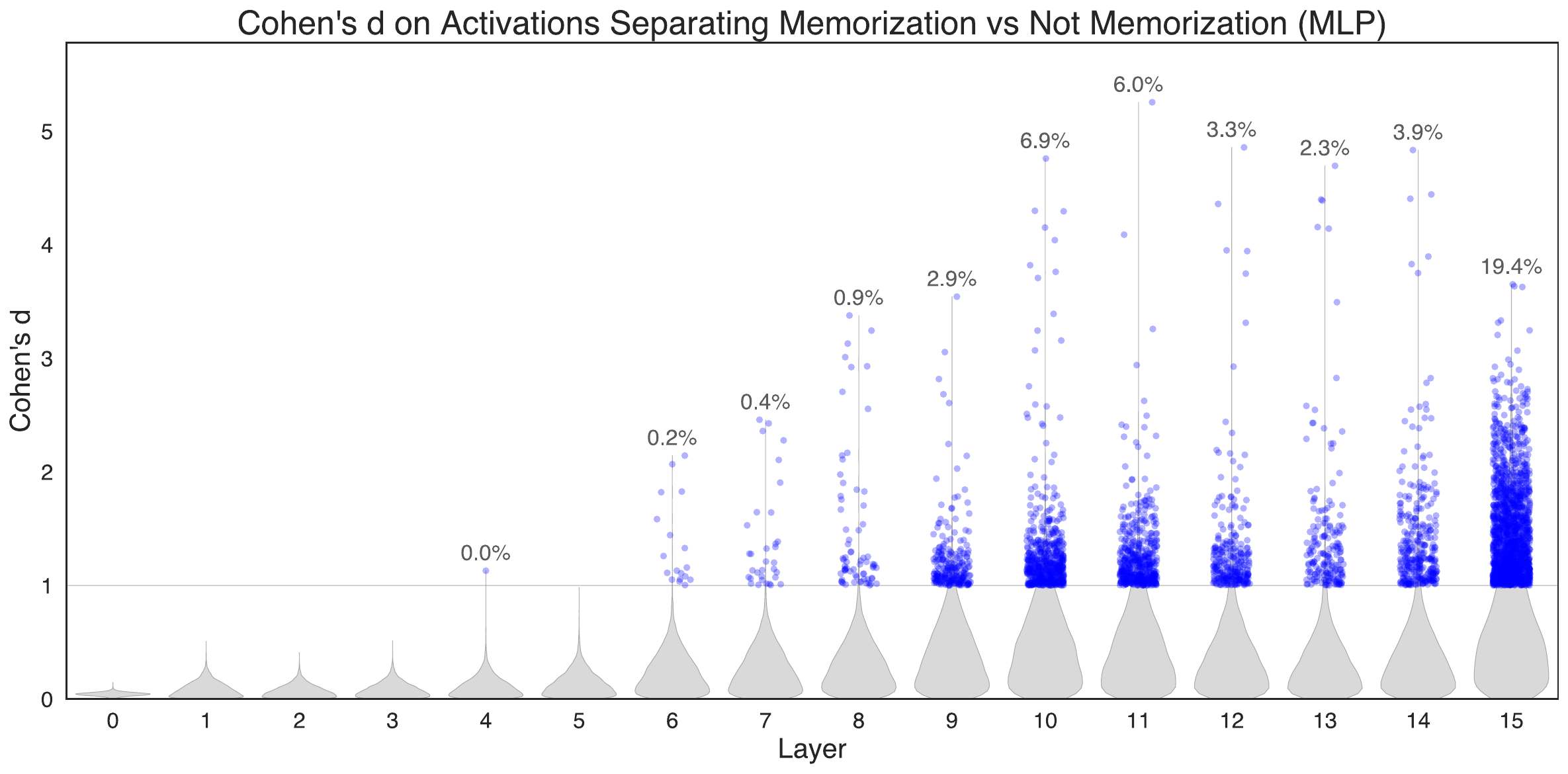}
\end{figure}

\begin{figure}[!htb]
    \centering
    \includegraphics[width=\textwidth]{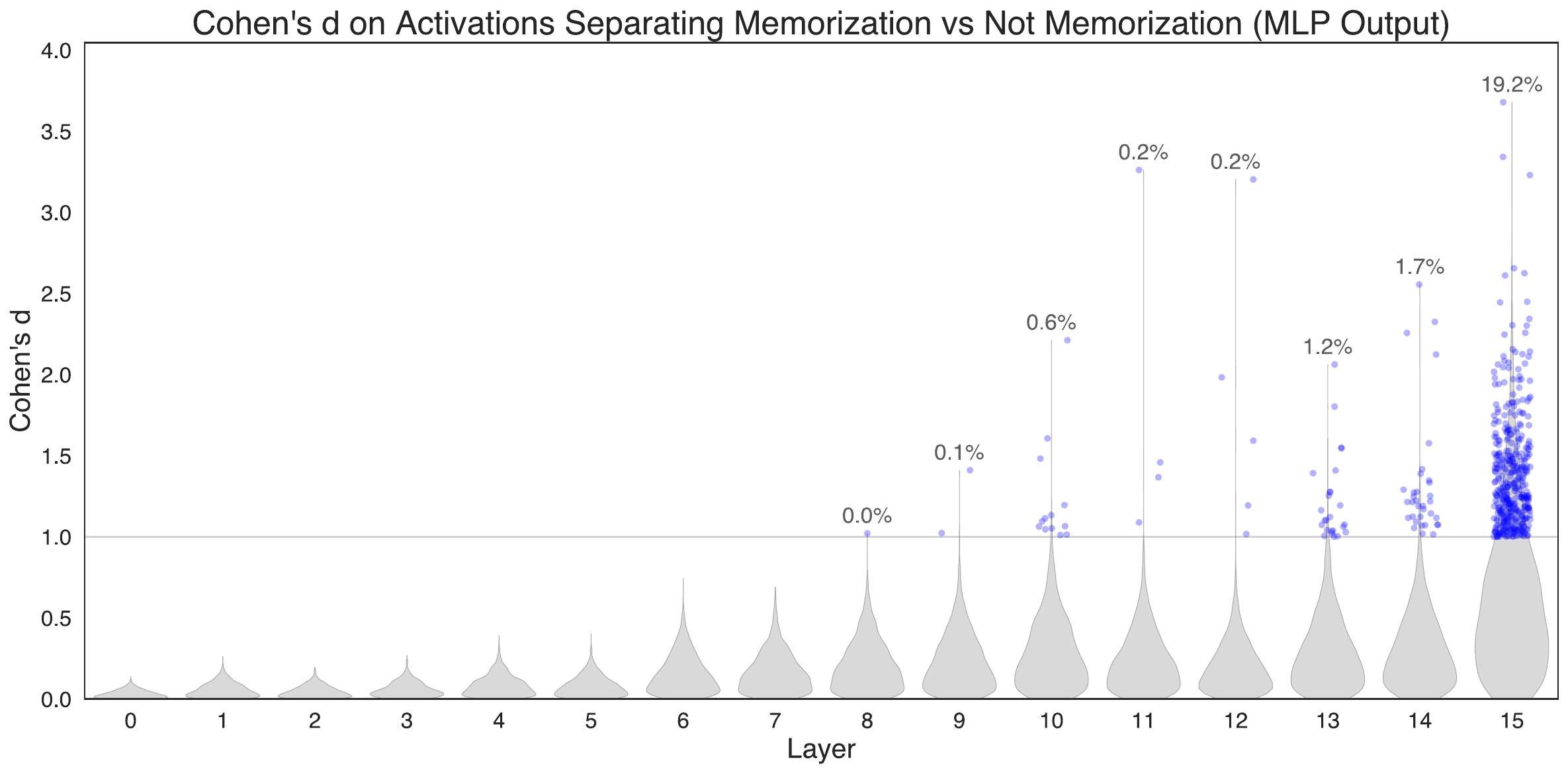}
\end{figure}

\begin{figure}[!htb]
    \centering
    \includegraphics[width=\textwidth]{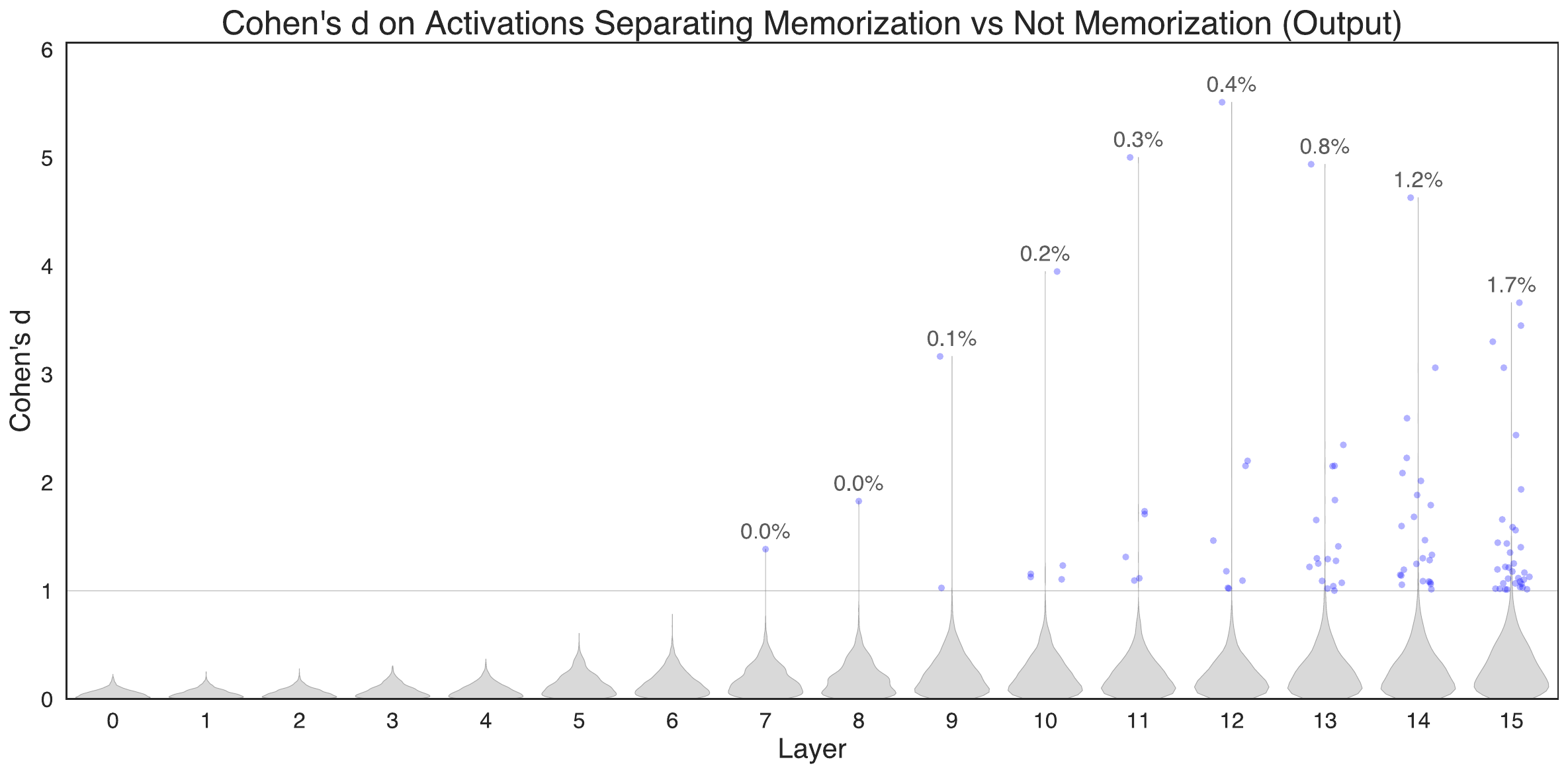}
\end{figure}

\section{Classification Accuracy Using Best Activation (Cohen's d) for Memorization}
\label{app:classification-accuracy-cohens-d}

\setlength{\intextsep}{5pt}
\setlength{\floatsep}{2pt}

\begin{figure}[!htb]
    \centering
    \includegraphics[width=\textwidth]{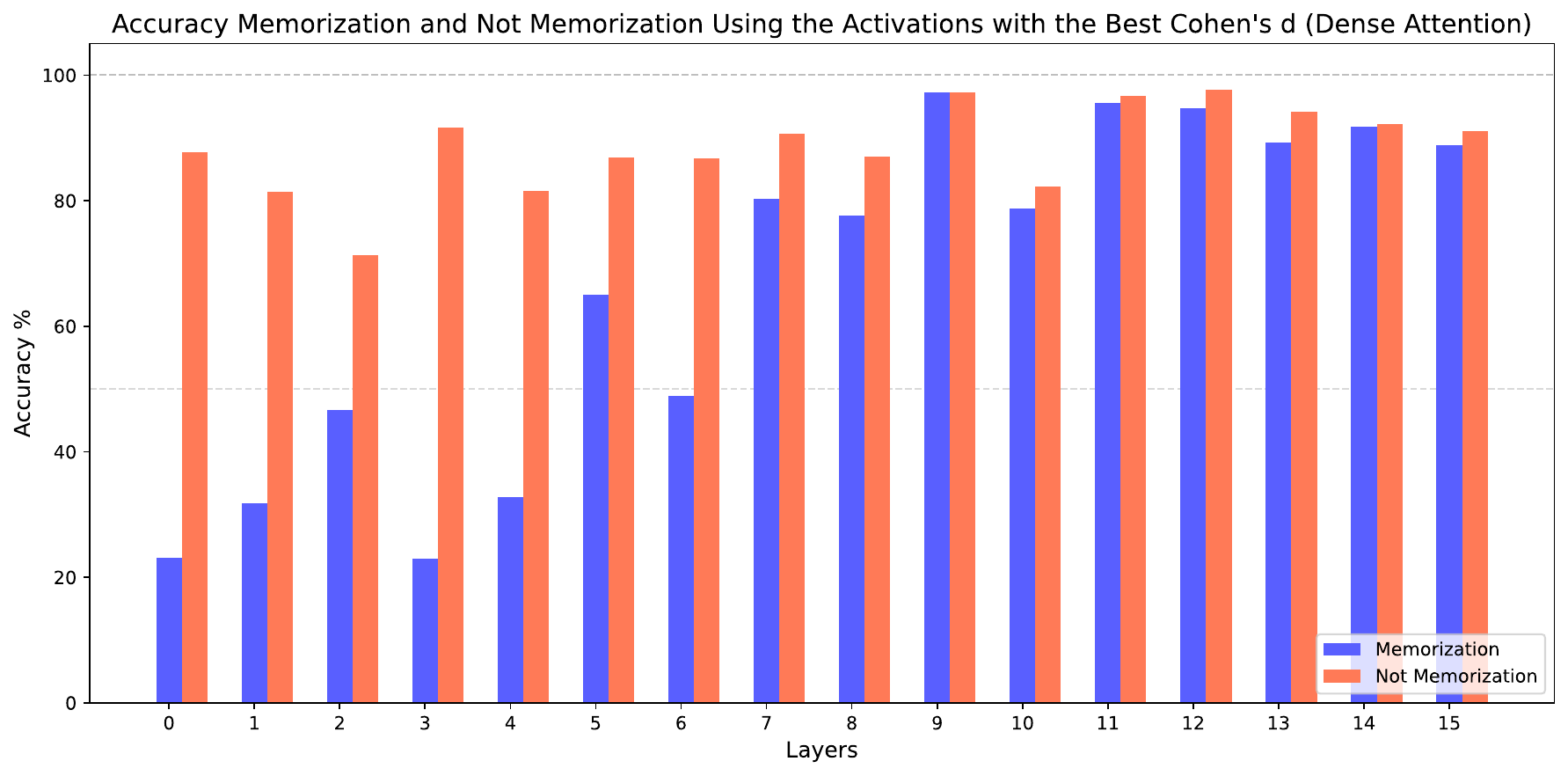}
\end{figure}

\begin{figure}[!htb]
    \centering
    \includegraphics[width=\textwidth]{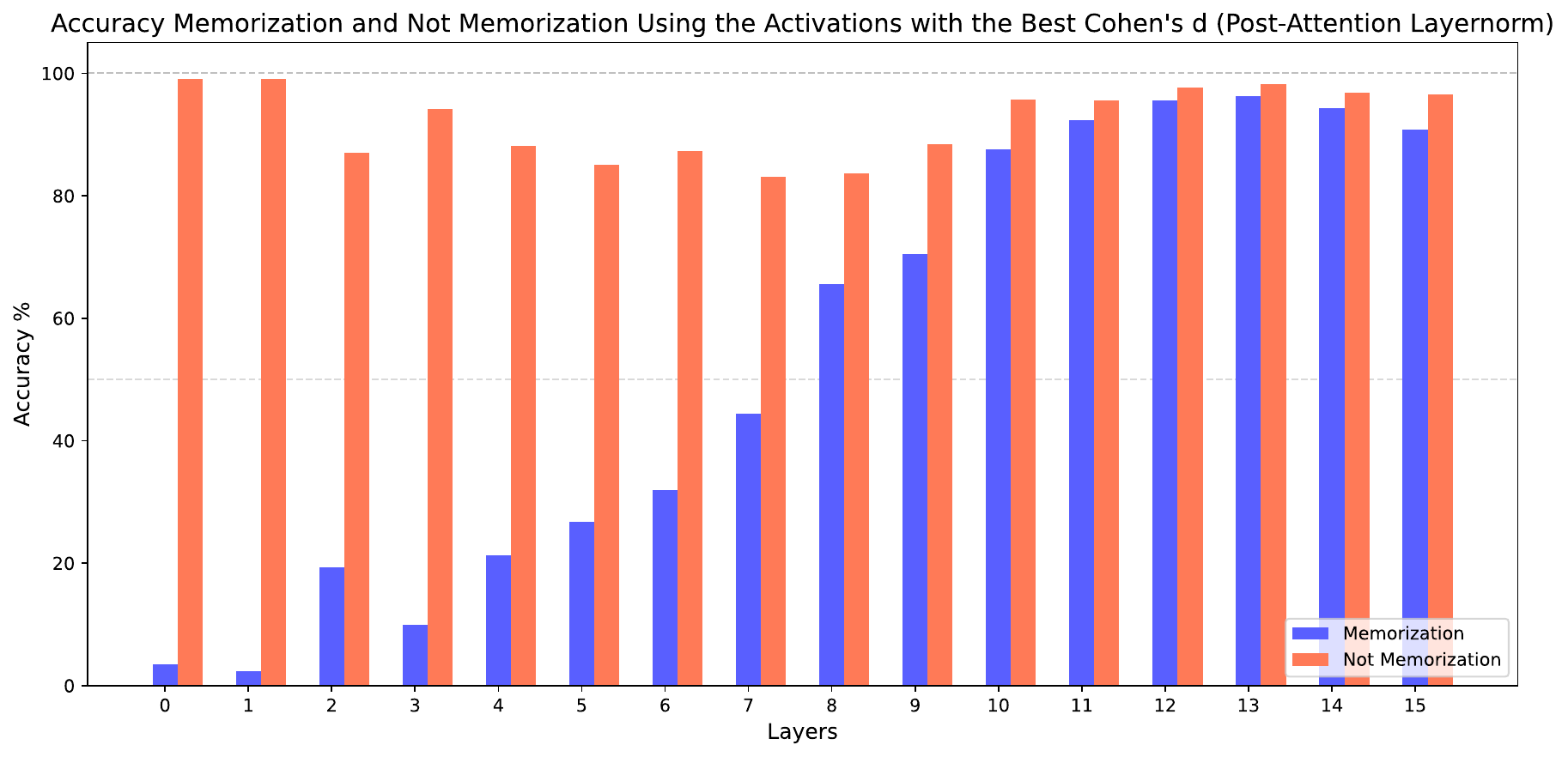}
\end{figure}

\begin{figure}[!htb]
    \centering
    \includegraphics[width=\textwidth]{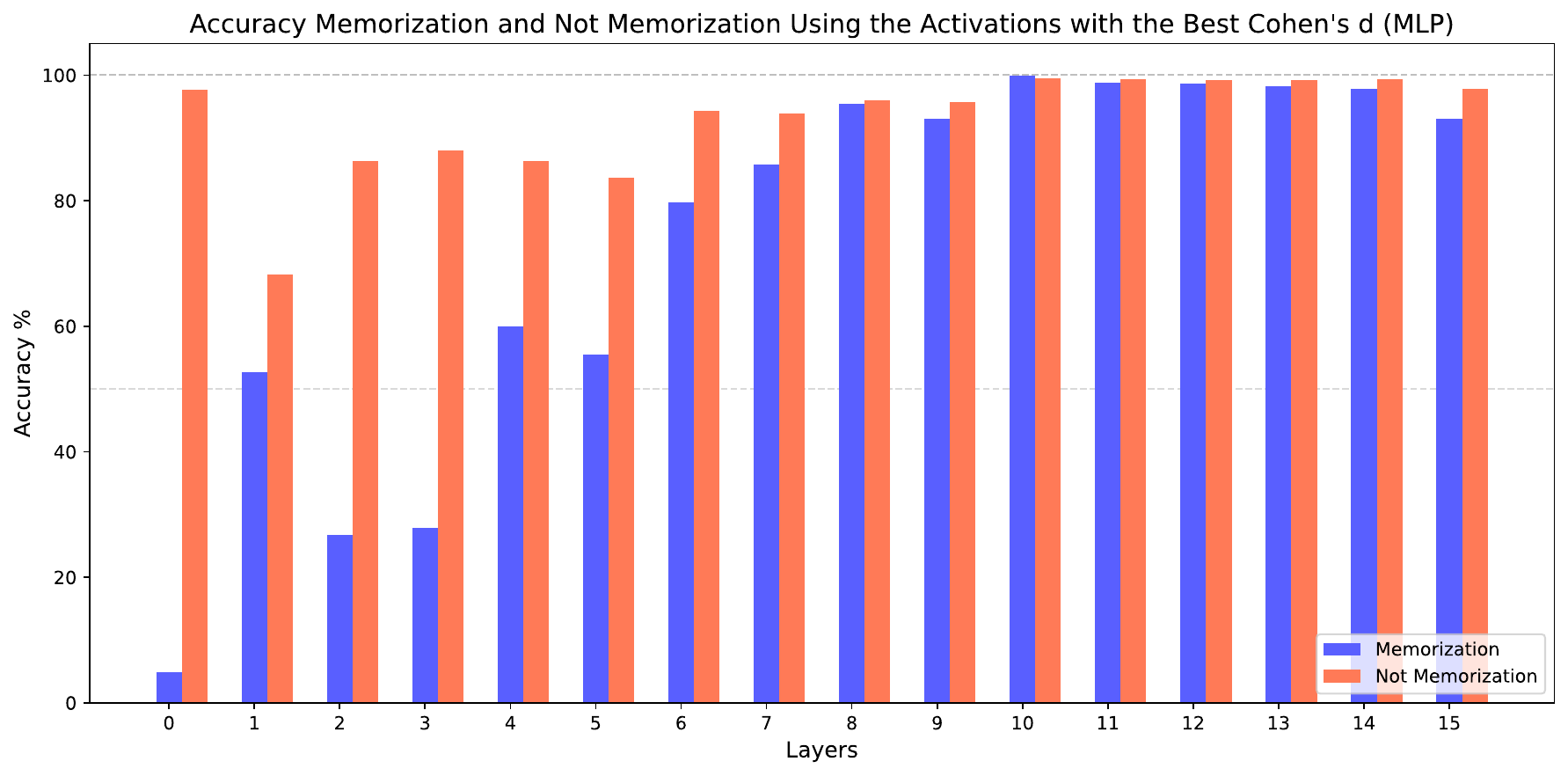}
\end{figure}

\begin{figure}[!htb]
    \centering
    \includegraphics[width=\textwidth]{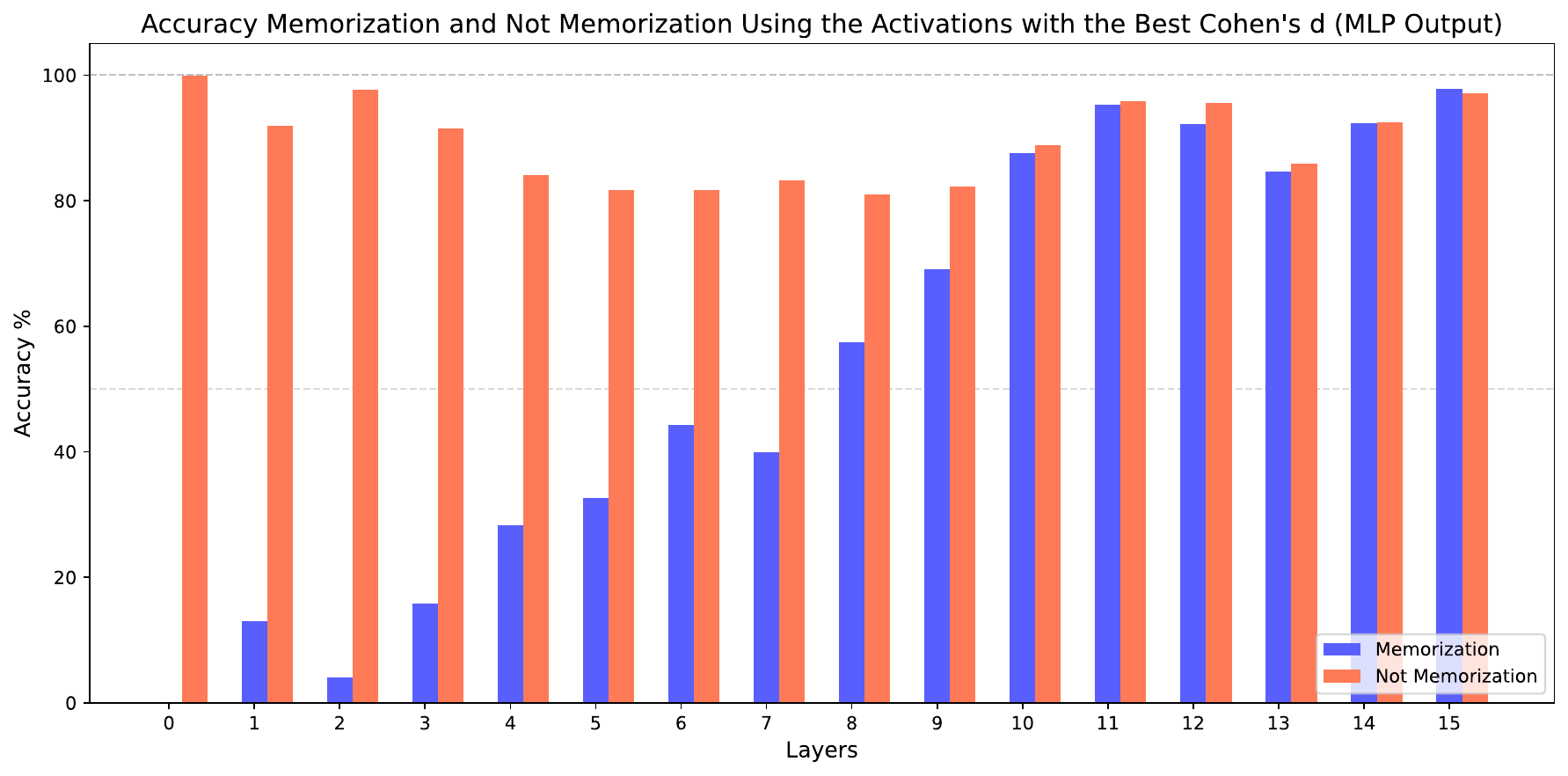}
\end{figure}

\clearpage
\begin{figure}[H]
    \centering
    \includegraphics[width=\textwidth]{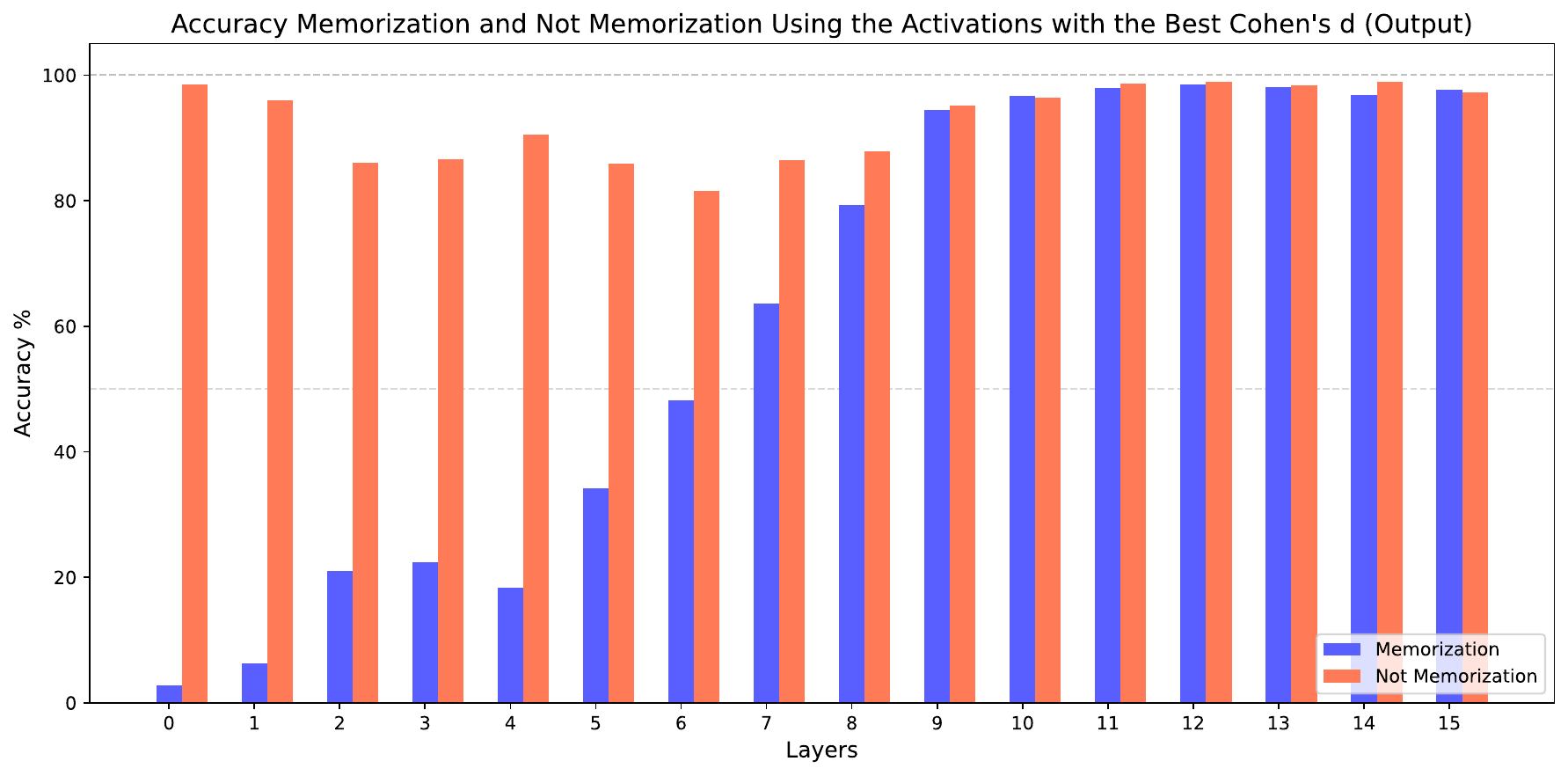}
\end{figure}

\end{document}